%% file: acl_latex.tex
\pdfoutput=1

\documentclass[11pt]{article}

\usepackage[final]{acl}

\usepackage{times}
\usepackage{latexsym}

\usepackage[T1]{fontenc}

\usepackage[utf8]{inputenc}

\usepackage{microtype}

\usepackage{inconsolata}

\usepackage{hyperref}
\usepackage{url}

\usepackage{amsmath}
\usepackage{amsfonts}
\usepackage{multirow}
\usepackage{booktabs,arydshln}
\usepackage{graphicx}
\usepackage{algorithm}
\usepackage{algorithmic}
\usepackage{tabu}
\usepackage{xcolor}
\usepackage{amssymb}
\usepackage{subfigure}
\usepackage{pdfpages}

\makeatletter
\def\adl@drawiv#1#2#3{%
        \hskip.5\tabcolsep
        \xleaders#3{#2.5\@tempdimb #1{1}#2.5\@tempdimb}%
                #2\z@ plus1fil minus1fil\relax
        \hskip.5\tabcolsep}
\newcommand{\cdashlinelr}[1]{%
  \noalign{\vskip\aboverulesep
           \global\let\@dashdrawstore\adl@draw
           \global\let\adl@draw\adl@drawiv}
  \cdashline{#1}
  \noalign{\global\let\adl@draw\@dashdrawstore
           \vskip\belowrulesep}}
\makeatother

\newcommand{\rom}[1]{\uppercase\expandafter{\romannumeral #1\relax}}

\title{X-Instruction: Aligning Language Model in Low-resource Languages with Self-curated Cross-lingual Instructions}

\author{Chong Li\footnotemark[2], Wen Yang\footnotemark[2], Jiajun Zhang\footnotemark[1], Jinliang Lu, Shaonan Wang, Chengqing Zong \\
        State Key Laboratory of Multimodal Artificial Intelligence Systems, \\
        Institute of Automation, CAS, Beijing, China\\
        School of Artificial Intelligence, University of Chinese Academy of Sciences, Beijing, China\\
        \{lichong2021, yangwen2023, lujinliang2019\}@ia.ac.cn,\\
        \{jjzhang, shaonan.wang, cqzong\}@nlpr.ia.ac.cn
        }

\begin{document}
\maketitle

\renewcommand{\thefootnote}{\fnsymbol{footnote}} 
\footnotetext[2]{These authors contributed equally to this work.}
\footnotetext[1]{Corresponding author.}

\renewcommand{\thefootnote}{\arabic{footnote}}

\begin{abstract}
Large language models respond well in high-resource languages like English but struggle in low-resource languages. 
It may arise from the lack of high-quality instruction following data in these languages. 
Directly translating English samples into these languages can be a solution but unreliable, leading to responses with translation errors and lacking language-specific or cultural knowledge. 
To address this issue, we propose a novel method to construct cross-lingual instruction following samples with instruction in English and response in low-resource languages. 
Specifically, the language model first learns to generate appropriate English instructions according to the natural web texts in other languages as responses. 
The candidate cross-lingual instruction tuning samples are further refined and diversified. 
We have employed this method to build a large-scale cross-lingual instruction tuning dataset on 10 languages, namely X-Instruction. 
The instruction data built using our method incorporate more language-specific knowledge compared with the naive translation method. 
Experimental results have shown that the response quality of the model tuned on X-Instruction greatly exceeds the model distilled from a powerful teacher model, reaching or even surpassing the ones of ChatGPT. 
In addition, we find that models tuned on cross-lingual instruction following samples can follow the instruction in the output language without further tuning. \footnote{Our code and data are available at \href{https://github.com/ZNLP/X-Instruction}{https://github.com/\\ZNLP/X-Instruction}}
\end{abstract}

\section{Introduction}
Large language models illustrate a series of remarkable abilities, e.g., generating better response and zero-shot task generalization, after tuning on instruction following samples \citep{long2022instructgpt, wei2022finetuned, mishra-etal-2022-cross}. 
However, the responses in low-resource languages, which are prone to contain unsafe content \citep{yong2023low} and ignore cultural nuances \citep{liu2023multilingual}, are inferior to the ones in high-resource languages. 
In addition to the limited pre-training corpus in these languages, it may come from the lack of high-quality instruction following data. 

\begin{figure}[t]
\centering
\subfigure[Instruction sample translated from English.]{\label{fig:bad_translate}\includegraphics [width=0.48\textwidth]{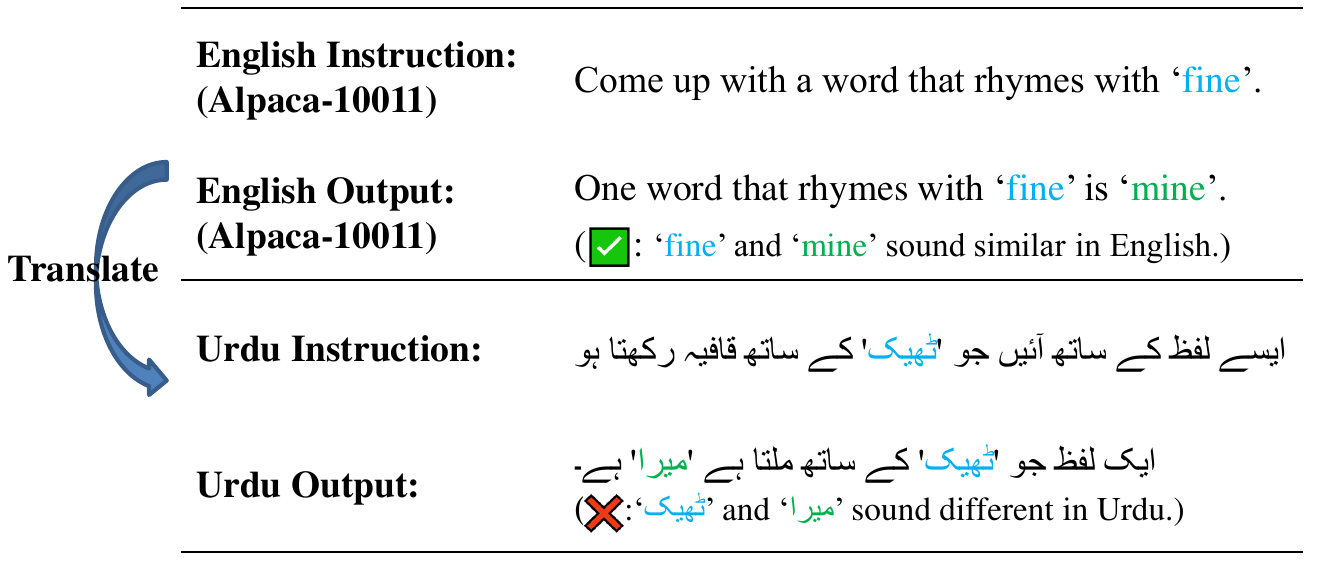}}
\subfigure[Cross-lingual instruction sample excavated.]{\label{fig:good_xIns}\includegraphics [width=0.48\textwidth]{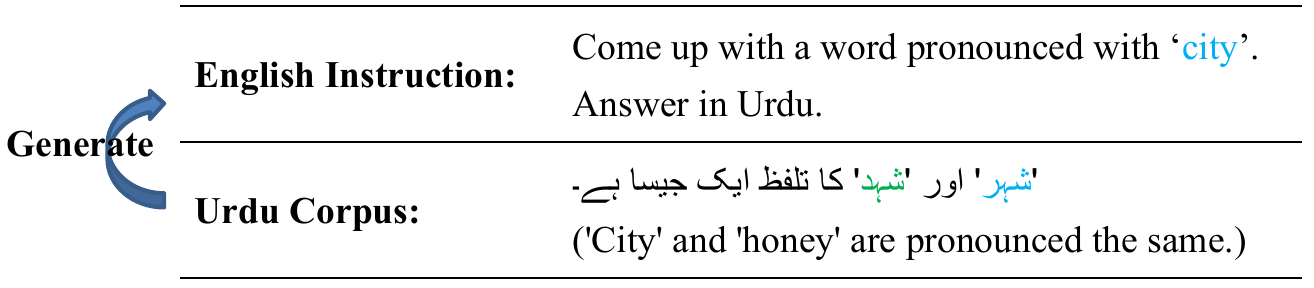}}
\vspace{-3mm}
\caption{(a) The instruction sample translated from English ignores the language-specific knowledge that ``fine'' and ``mine'' sound different in Urdu. (b) The cross-lingual instruction tuning sample generated by our method contains language-specific knowledge from the native corpus. }
\vspace{-3mm}
\end{figure}

Distillation from the teacher model or manual annotation are common methods to obtain high-quality instructions. 
The former cannot be applied to low-resource languages since the teacher model may respond poorly in these languages. 
Considering the high requirements, including creative thinking and professional knowledge, to write instruction-following samples, it is hard to find suitable annotators that excel in low-resource languages. 
Translating English instruction tuning samples into the target language is also unreliable, which will introduce translation errors, especially in the samples with codes, and overlook the cultural nuances between languages.  
As shown in \hyperref[fig:bad_translate]{Figure 1(a)}, the Urdu sample translated is wrong due to the pronunciation nuance between English and Urdu. 
Therefore, aligning language models in low-resource languages remains a significant challenge. 

In the preliminary experiment, we found that LLaMA \citep{touvron2023llama, touvron2023llama2} can understand unseen low-resource language instructions quite well through responding in English, when tuned with limited instruction-following samples, but it performs much worse when replying in its own language (Table \ref{tab:preliminary}). 

Based on the observation, we first adopt the cross-lingual instruction following format like Figure \ref{fig:good_xIns}, in which the languages of instructions and responses are not the same, to exploit the understanding ability rapidly learned in low-resource languages and better generation performance in English. 
We further propose an automatic pipeline to excavate and refine cross-lingual instruction tuning data from the unsupervised multilingual corpus, which contains language-specific knowledge and native expressions. 
Specifically, the model first learns from seed data to generate appropriate English instructions for the given texts in other languages, which compose candidate cross-lingual instruction following samples. 
The candidate samples are refined in an iterative manner. 
In each iteration, an evaluator is trained through a synthetic rating dataset and finds better cross-lingual samples, which are exploited to improve the evaluator in the next iteration. 
Thus, the quality of samples found is improved with the stronger evaluator. 
The final cross-lingual instruction tuning dataset is sampled from the different clusters of the remaining data after $k$-th iteration to increase the diversity. 

We conducted experiments on 10 languages, which include 5 medium-resource languages and 5 low-resource languages\footnote{We follow the language category used in \citet{lin-etal-2022-shot} sorted by the data ratios in the CC100 corpus \citep{conneau-etal-2020-unsupervised, wenzek-etal-2020-ccnet}.}, and generated 320k high-quality cross-lingual instruction tuning samples. 
On four instruction-following evaluation benchmarks, the response quality of the base model tuning on our data can surpass those generated from translation and distillation baseline models, and even exceed the ones of ChatGPT \citep{openai2022chatgpt} in these languages. 
In addition, we find that cross-lingual instruction tuning models can follow the instruction in the language of output without further tuning and maintain 90.6\% response quality. 

To sum up, our contributions are as follows:
\begin{itemize}
    \item We propose an automatic pipeline to excavate cross-lingual instruction-tuning samples with language-specific knowledge. 
    \item We have built and will release a high-quality cross-lingual instruction-tuning dataset for 10 languages, including 5 low-resource languages. Experimental results demonstrate that models tuning on it can generate better responses in low-resource languages. 
    \item We find that cross-lingual instruction tuning models obtain a zero-shot instruction following ability in the output language. 
\end{itemize}

\input{tabs/sw_ur_gpt-4}

\section{Related Works}
Our work is related to multilingual instruction generation and tuning, which will be briefly introduced below. 
\paragraph{Instruction Generation}
Large language models obtain instruction following and better zero-shot task generalization ability after tuning on diverse instruction following samples \citep{wei2022finetuned, mishra-etal-2022-cross, chung2022scaling}. 
The key challenge is how to build high-quality instruction following samples that have a great influence on the performance of the model after tuning. 
Distilling from more powerful models \citep{wang-etal-2023-self-instruct, taori2023alpaca, xu2023wizardlm} and human labeling \citep{kopf2023openassistant, zhou2023lima} are main methods to generate high-quality instruction following samples. 
Our work is more similar to the self-align methods \citep{bai2022constitutional, sun2023principle, li2023self}, which show promising results by iteratively self-generating and filtering instructions without strong models and a lot of manual annotation. 
However, it cannot be applied to the condition when language models are unable to generate high-quality instructions in unseen languages. 
In contrast, our method avoids the weakness in the low-resource language generation and exploits superior generation ability in English. 

\begin{figure*}[t]
\centering 
\setlength{\textfloatsep}{2pt}
\setlength{\intextsep}{2pt}
\setlength{\abovecaptionskip}{2pt}
\includegraphics[width=1\textwidth]{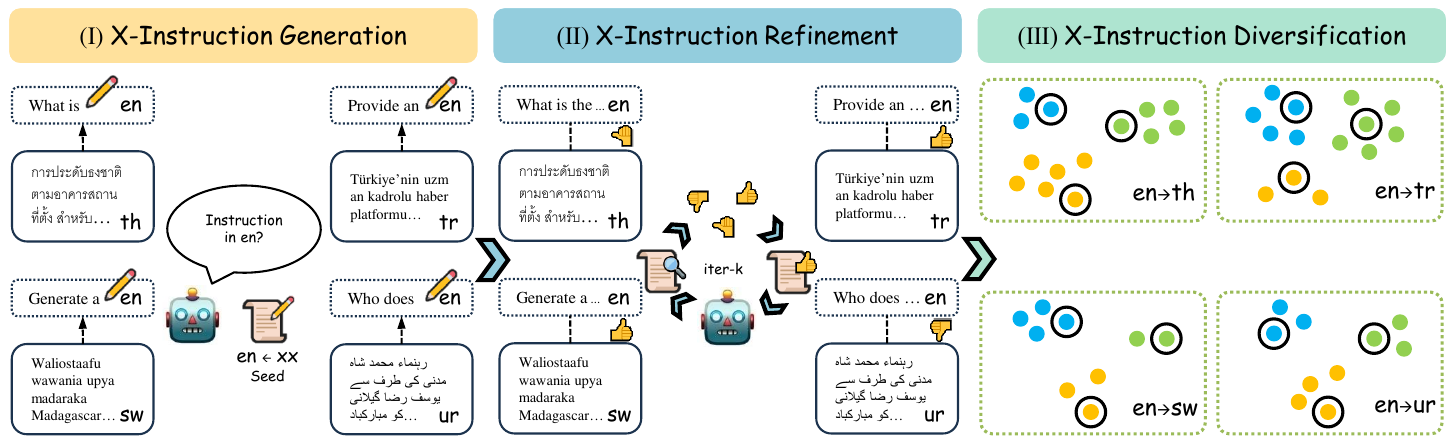}
\caption{Illustration of how to generate and refine cross-lingual instruction (X-Instruction) examples. (\rom{1}) Language models learn to generate cross-lingual instructions for multilingual texts using seed data. (\rom{2}) Language models iteratively label and refine cross-lingual instruction samples. (\rom{3}) The final instruction data are sampled from different clusters of embedding from the English instruction to increase the diversity. 
}\label{fig:framework}
\vspace{-4mm}
\end{figure*}

\paragraph{Multilingual Instruction Tuning}
After tuning on large-scale multilingual instruction following samples translated from English, large language models show better zero-shot multilingual performance and language generalization results \citep{muennighoff-etal-2023-crosslingual, li2023align}. 
On the other hand, \citet{li2023bactrian} translate 67k instructions only and distill language-specific outputs from ChatGPT, which shows better responses and multilingual abilities than the ones tuning on translated samples. 

Different from distilling language-specific outputs from more powerful models, our method attempts to excavate high-quality samples from the neural multilingual corpus and is applicable to languages that are not supported by teacher models. 

\section{Method}
As shown in Figure \ref{fig:framework}, we design a three-step pipeline to automatically excavate high-quality cross-lingual instruction following samples from multilingual web corpora: 
1) The base language model is fine-tuned to generate candidate English instructions given multilingual corpus (Section \ref{sec:xIns_generation}). 
2) Samples with inappropriate cross-lingual instruction and output are discarded to improve the quality of the dataset (Section \ref{sec:xIns_refinement}). 
3) We diversify the cross-lingual instruction data by sampling from the multiple clusters of the English instruction embedding (Section \ref{sec:xIns_diversification}).

\subsection{X-Instruction Definition}
The instruction following sample consists of an instruction $i$, an optional input $x$, and an output $y$ in the same language. 
Considering the vague boundary between instruction and input, the input $x$ is ignored in this work for simplicity. 
We define the \textbf{Cross}-lingual \textbf{Instruction} (\textbf{X-Instruction}) sample $(i^{a}, y^{b})$ that the language $a$ of instruction is different from the one of output ($b$). 
Taking the cross-lingual instruction sample in \hyperref[fig:bad_translate]{Figure 1(b)} as an example, the language of instruction is English, while the language of output is Urdu. 

\subsection{X-Instruction Generation}
\label{sec:xIns_generation}
To exploit the better generation performance in high-resource languages like English, we generate English instructions for a corpus in other languages. 
Specifically, given seed cross-lingual instruction samples $\mathcal{D}_0=\{(i_j^{h}, y_j^{l})\}_{j=1}^{n_s}$, where the $h$ and $l$ denote the high- and low-resource languages respectively, the language model is fine-tuned to generate the instruction $i_j^h$ given the response $y_j^{l}$ as input.
Then, we use it to generate candidate cross-lingual instructions for the multilingual corpus, basing the assumption that some texts among the corpus are good responses in X-Instruction samples. 

\subsection{X-Instruction Refinement}
\label{sec:xIns_refinement}
After generating cross-lingual instruction samples, it is important to find and discard the inappropriate ones due to the quality of instruction following samples having a great influence on the performance of model aligned \citep{chen2023alpagasus, zhou2023lima}. 
To achieve this, we design an iterative method named ``X-Instruction Refinement'', which is shown in \hyperref[fig:framework]{Figure 2(\rom{2})}. 

In the $k$-th iteration, we first synthesize a pseudo-rating dataset $\mathcal{D}_k^r=\{(i_j^{h}, \hat{y}_j, s_j)\}_{j=1}^{n_r}$ from a part of seed data $\mathcal{D}_0^r$, where $\mathcal{D}_0^r \subset \mathcal{D}_0$, to train the evaluator that outputs three-level ratings. 
Given the instruction $i_j^{h}$, the best response ($s_j=2$) is the vanilla output $y_j^{l}$, while the worst one ($s_j=0$) is selected from the mismatched output $y_m^{l}$ ($m \neq j$). 
The reasonable but flawed output ($s_j=1$) is chosen from the modified output $\hat{y}_j^{l}$ with some parts deleted or duplicated in the vanilla output $y_j^{l}$, or the output of the cross-lingual instruction-following model tuned on $\mathcal{D}^{x}_{k-1}$. 
In the first iteration ($k=1$), the seed data $\mathcal{D}^{x}_{0}=\mathcal{D}_0\setminus \mathcal{D}_0^r$, which is not used in the pseudo-rating dataset $\mathcal{D}_0^r$, is adopted to fine-tune the cross-lingual instruction following model. 
In the subsequent iteration ($k\geq$2), $\mathcal{D}^{x}_{k-1}$ consists of the best cross-lingual instruction samples found by the evaluator and $\mathcal{D}^{x}_{0}$. 

As the iteration proceeds, the quality of output from the instruction following model improves by tuning on higher quality X-Instruction samples, which reduces the gap between the second-level sample and the highest one in the pseudo-rating dataset. 
Thus, the trained evaluator can find better cross-lingual instruction tuning samples. 
The number of iterations is set to 3 by default, which is investigated in Section \ref{sec:analyse_iter}. 

\subsection{X-Instruction Diversification}
\label{sec:xIns_diversification}
In addition to the quality, the diversity of instruction tuning samples also has a great impact on the performance \citep{wan-etal-2023-explore}. 
To diversify X-Instruction samples, we first obtain the embedding for each instruction $i^{h}$ using a pre-trained sentence encoder. 
After applying \textit{k}-means on these embeddings, the same amount of X-Instruction samples are sampled from each cluster (\hyperref[fig:framework]{Figure 2(\rom{3})}). 
It aims to diversify cross-lingual instruction examples by avoiding the dominance of samples from a few domains in the final dataset.

\section{X-Instruction Dataset}
We apply the automatic pipeline to a large-scale multilingual corpus using LLaMA-2-7B, and construct a cross-lingual instruction tuning dataset with 320k samples for 10 languages, which is named \textbf{X-Instruction dataset}. 
Codes and data will be made public after review to advocate future research. 

\subsection{Seed Data}
For each language, we extract the first conversation turn of message trees in the Open Assistant dataset, and adopt the highest quality response as the output for each sample \citep{kopf2023openassistant}. 
To construct the X-Instruction sample, the instruction of each sample is translated into English by Google Translate. 
However, there are only a few or even no human-labeled samples for the most of languages involved in this work. 
Thus, we translate the output of 3k English samples into the target language using Google Translate to supplement the seed data for each language. The statistics of seed data are reported in Appendix \ref{appendix:seed_data}. 

\subsection{Unsupervised Multilingual Corpus}
The multilingual corpus for each language is extracted from the CulturaX dataset \citep{nguyen2023culturax}, which contains web texts in 167 languages and is filtered from mC4 \citep{xue-etal-2021-mt5} and OSCAR \citep{suarez2019asynchronous}. 
We only sample 1M web texts for each language from this dataset due to the constraint on the computation budget. 

\subsection{Statistics}
To investigate the diversity of X-Instruction, we parse English instructions and count the ones with verb-noun structure using the Berkeley Neural Parser \citep{kitaev-klein-2018-constituency, kitaev-etal-2019-multilingual}. 
Figure \ref{fig:verb_noun} illustrates the top 16 most common root verbs and their top direct noun objects. 
We can find that the great diversity of X-Instruction constructed from web corpora. 
The additional information of X-Instruction is reported in Table \ref{tab:statis}.

\begin{figure}[th]
\centering
\includegraphics[width=0.45\textwidth]{imgs/xIns_pile.pdf}\vspace{-2mm}
\caption{\label{fig:verb_noun} The Statistic of the top 16 verbs (inner circle) and their top direct nouns (outer circle) in English instructions from X-Instruction.}
\vspace{-4mm}
\end{figure}

\input{tabs/xins_statistic}

\subsection{Quality}
For each language, we randomly sample 200 instructions and corresponding texts to further evaluate the quality. 
There are two questions designed to conduct this evaluation, and results are reported in Table \ref{tab:quality_eval}. 
It can be found that the quality of X-Instruction is good for more than 80\% of samples are valid. 
The top-flows come from additional information in web text like navigation bar (11.3\%), incorrect URL and HTTP status codes (3.8\%). 

\input{tabs/quality_evaluation}

\section{Experiments}
\subsection{Experiments Settings}
\paragraph{Base models} LLaMA-2 models with 7B and 13B parameters are taken as base models and fine-tuned on cross-lingual samples in each language from the X-Instruction dataset. 
Hyperparameters are reported in Appendix \ref{appendix:param}. 

\paragraph{Baselines} used in this work are:
\begin{itemize}
\itemindent=-5pt
\item \textbf{Alpaca-MT}: We translate the vanilla Alpaca dataset \citep{taori2023alpaca} into 10 languages using Google Translate and fine-tune the same base model for comparison. 

\item \textbf{$\text{Bactrian-X}$} \citep{li2023bactrian}: LLaMA models are fine-tuned with 3.48M instruction tuning samples in the Bactrian-X dataset, which consists of outputs from ChatGPT in 51 languages except English. 

\item \textbf{$\text{Bactrian-M}$}: We fine-tune LLaMA-2 models using 67k monolingual instruction tuning samples in each language from the Bactrian-X dataset. It distills the monolingual capabilities of ChatGPT, providing a strong baseline for our work. 

\item \textbf{ChatGPT} \citep{openai2022chatgpt} is fine-tuned through reinforcement learning with human feedback\citep{ouyang2022training} on the GPT-3.5 model. It demonstrates remarkable multilingual capabilities across a wide range of tasks \citep{asai2023buffet}. We use the ``gpt-3.5-turbo-0613'' API in this work. 
\end{itemize}

\paragraph{Datasets} 
To confirm the effectiveness of X-Instruction, we conduct the evaluation of open-end generation on Vicuna \citep{zheng2023judging}, WizardLM \citep{xu2023wizardlm}, LIMA \citep{zhou2023lima}, and Koala \citep{Geng2023koala} datasets, which cover a variety of task categories. 
Prompts in these datasets are translated into the 10 languages involved. 
A comprehensive overview of these datasets is presented in Appendix \ref{appendix:dataset}. 

\input{tabs/four_benchmark_win_rate_on_three_language_v2}
\input{tabs/two_benchmark_on_ten_languages}

\begin{figure*}[ht]
\centering
\subfigure[${\text{X-Instruction}}_{\text{13B}}$ vs. ${\text{Bactrian-M}}_{\text{13B}}$]{
    \label{fig:radar_X-Instruct_vs_bx-s}
    \includegraphics[width=0.31\textwidth]{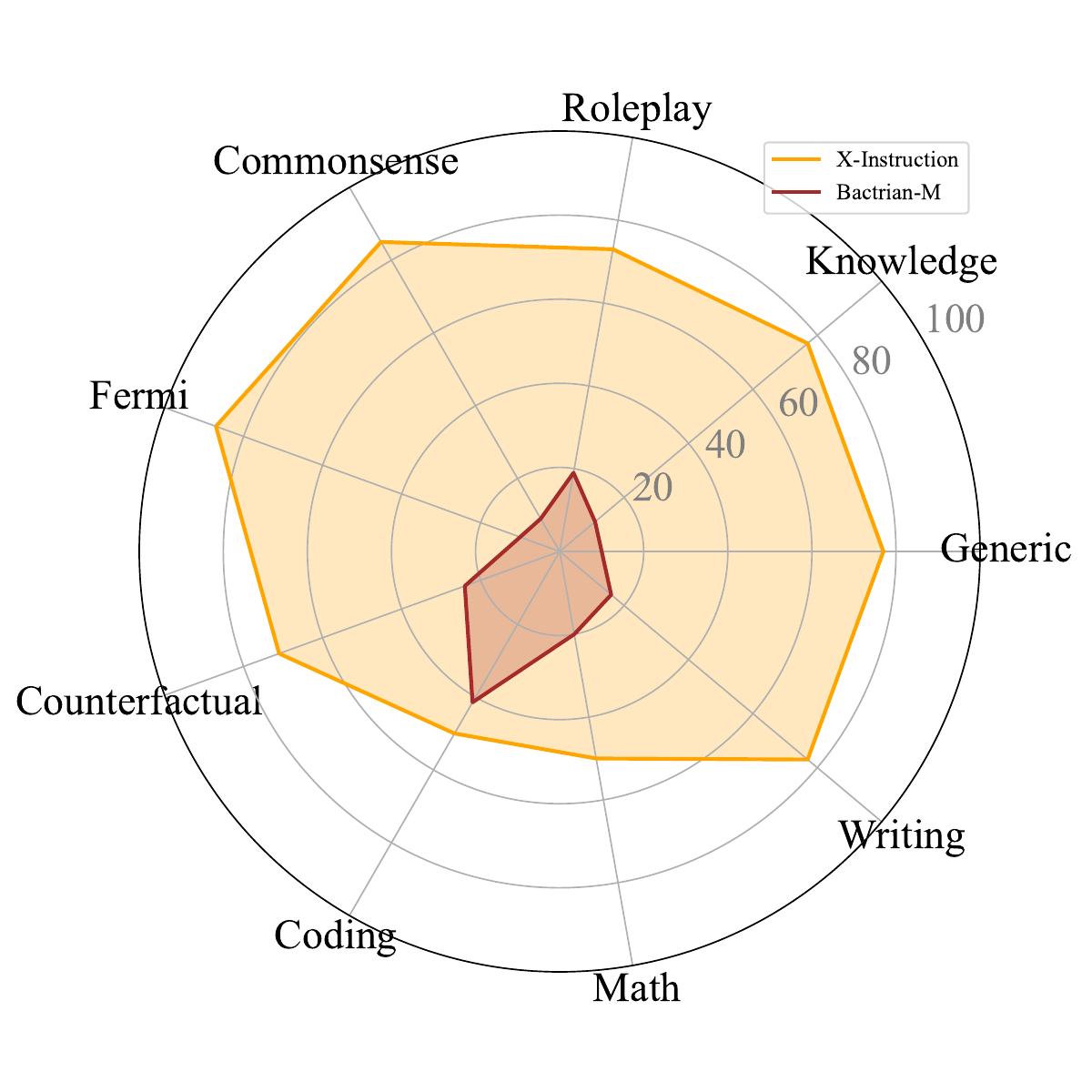}
}
\subfigure[${\text{X-Instruction}}_{\text{13B}}$ vs. ChatGPT]{
    \label{fig:radar_X-Instruct_vs_ChatGPT}
    \includegraphics[width=0.31\textwidth]{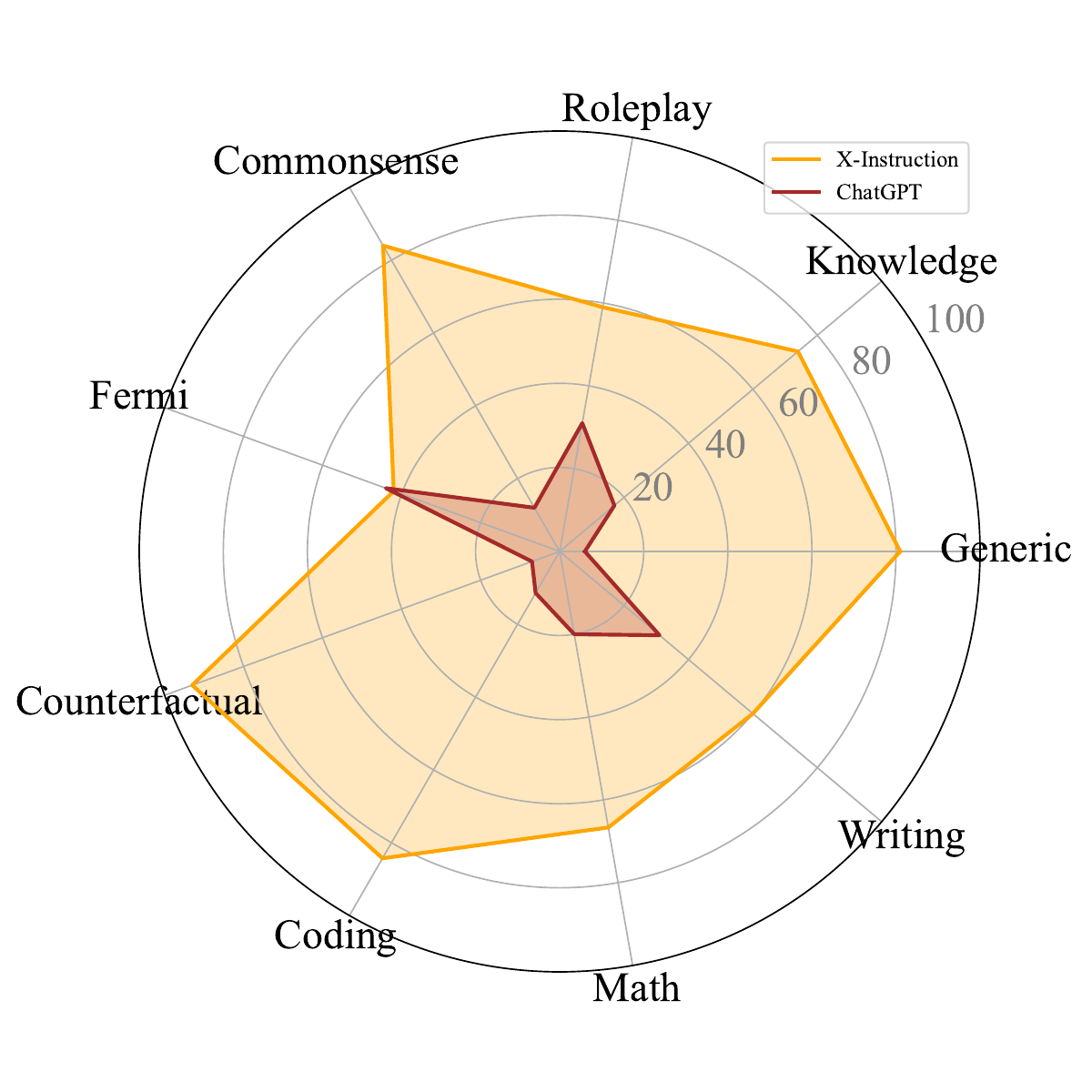}
}
\subfigure[${\text{X-Instruction}}_{\text{13B}}$ vs. ${\text{X-Instruction}}_{\text{13B}}^{\dagger}$]{
    \label{fig:radar_X-Instruct_vs_X-Instruct_Zero_shot}
    \includegraphics[width=0.31\textwidth]{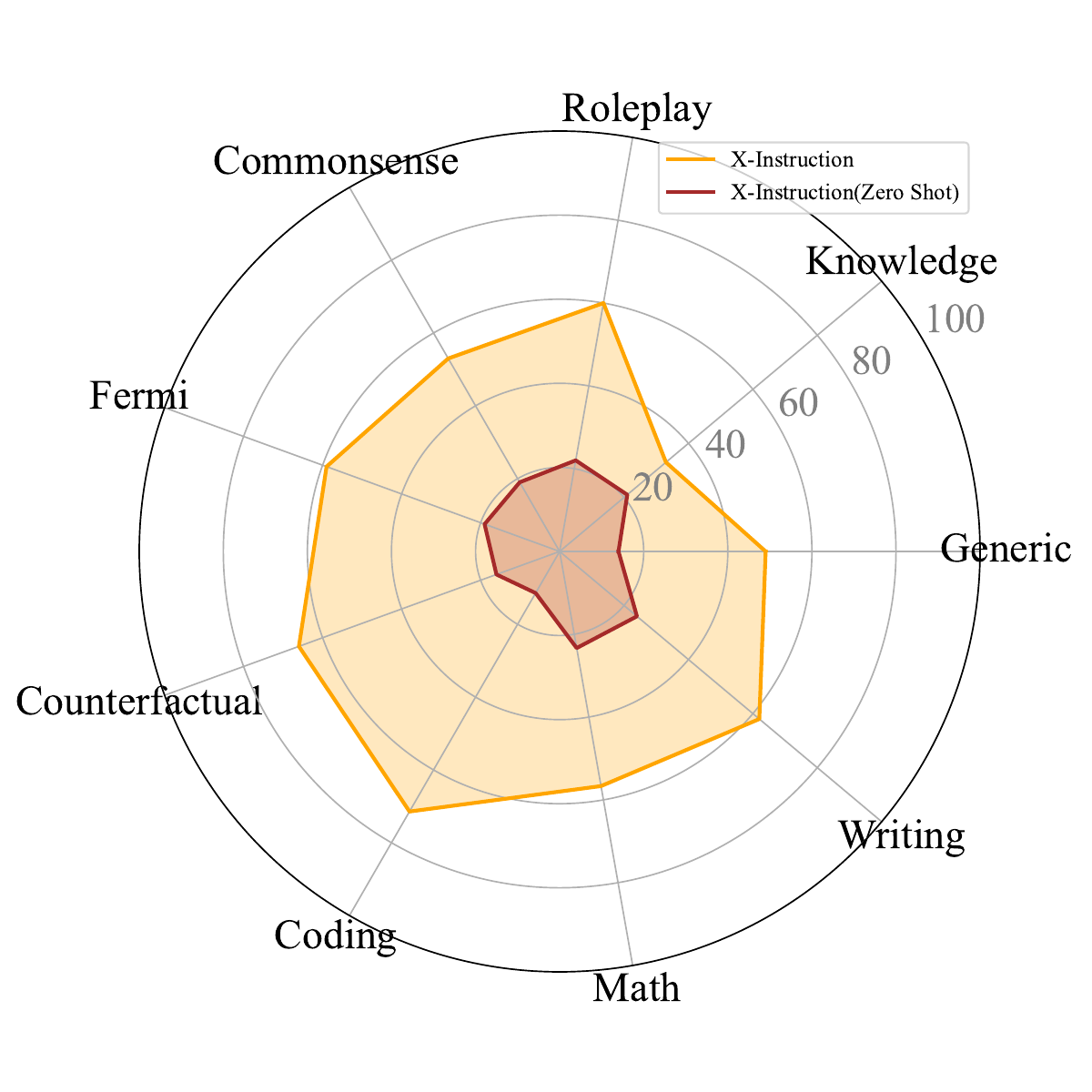}
}
\vspace{-3mm}
\caption{The average win rates distribution of $\text{X-Instruction}_{\text{13B}}$ vs. $\text{Bactrian-M}_{\text{13B}}$, ChatGPT and ${\text{X-Instruction}}_{\text{13B}}^{\dagger}$ on Vicuna dataset in 10 languages, where ${}^{\dagger}$ denotes the zero-shot evaluation results.}
\label{fig:radar_Vicuna}
\vspace{-2mm}
\end{figure*}

\subsection{Evaluation Results from GPT-4}
We take GPT-4 \citep{openai2023gpt4}\footnote{It points to the ``gpt-4-0613'' API during our experiments.} as a judge to conduct automatic evaluation, which is found a higher correlation with human judgements \citep{liu-etal-2023-g, alpaca_eval}.
Considering the excellent multilingual understanding ability of GPT-4 \citep{openai2023gpt4}, it is reasonable and effective to employ GPT-4 to automatically evaluate the quality of responses in low-resource languages. 

Specifically, we adopt pair-wise evaluation and request GPT-4 to determine the better response between responses $(r_1, r_2)$ from different models given the instruction $i$. 
In the evaluation, GPT-4 judge outputs a score, named GPT-4 score in this work, from 0 to 10 based on their helpfulness, relevance, and accuracy. 
The details of GPT-4 evaluation prompt can be found in Appendix \ref{appendix:gpt4_template}.

To alleviate the position bias in the evaluation by GPT-4 \citep{zheng2023judging}, we first request GPT-4 to evaluate $(r_1, r_2)$, then switch the position of $r_1$ and $r_2$, which is $(r_2, r_1)$, in the second evaluation. 
The better response is the one that wins twice or wins once and draws once. 

Table \ref{tab:win_rates_all_benchmarks} reports the win rates of models against ChatGPT on four benchmarks in three low-resource languages.  
It can be found that X-Instruction exhibits a significant performance advantage over Alpaca-MT and Bactrian-X in all benchmarks. 
The average improvement of X-Instruction models over Bactrian-M models, which distill outputs from ChatGPT, reaches 15.8\%. 
Compared with ChatGPT, $\text{X-Instruction}_{\text{13B}}$ demonstrates even better performance with a 67.66\% win rate. 

\begin{figure*}[ht]
\centering 
\setlength{\textfloatsep}{2pt}
\setlength{\intextsep}{2pt}
\setlength{\abovecaptionskip}{2pt}
\includegraphics[width=1\textwidth]{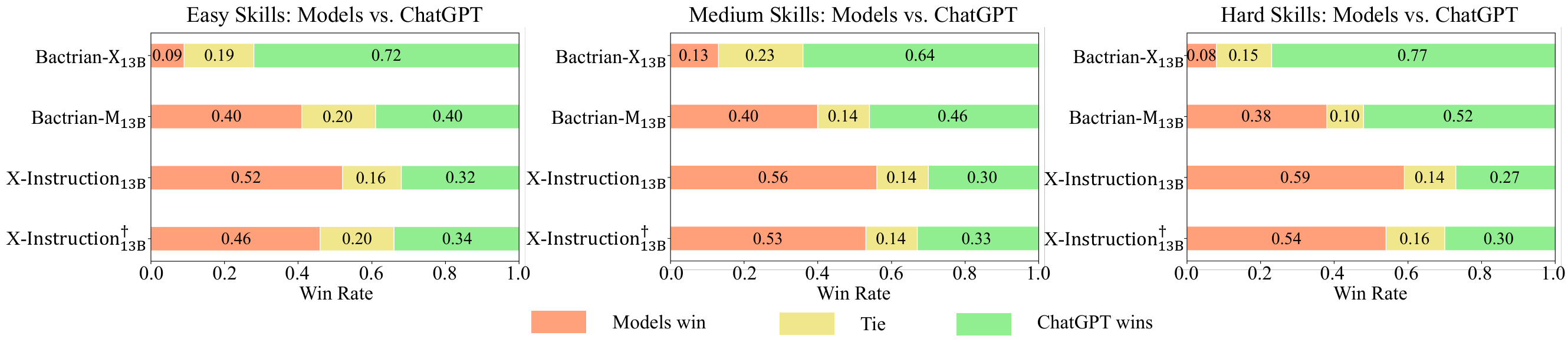}
\caption{The performance of models given prompts with different difficulties on WizardLM dataset in 10 languages. ${}^{\dagger}$ indicates the X-Instruction model is prompted in the output language under the zero-shot evaluation.}
\label{fig:difficulty_analysis}
\vspace{-4mm}
\end{figure*}

\paragraph{Languages Generalization}
To evaluate the effectiveness of X-Instruction in diverse languages, we extend experiments to ten languages, including five medium-resource and five low-resource languages, on Vicuna and WizardLM datasets. 
Table \ref{tab:win_rates_scores_all_langs} reports the detailed results on ten languages across two benchmarks, which exhibit a certain level of uniformity across ten languages. 
The X-Instruction models with only 3k labeled seed data on each language obtain the highest win rate across ten languages on average, indicating that our method can be extended to more languages. 

We further examine the generative quality by calculating the average GPT-4 score for each model from all comparison pairs, shown in Table \ref{tab:win_rates_scores_all_langs}. 
The average generative quality of X-Instruction is the most pronounced, achieving an average score of 6.9 in all languages. 
Notably, X-Instruction outperforms Bactrian-X by a large margin (1.3) and achieves better performance than Bactrian-M (5.6). 
Although the response quality of $\text{X-Instruction}_{\text{13B}}$ drops to 7.34 in five low-resource languages, the average win rate against ChatGPT increased by 14.9\% compared to the one in the other 5 medium-resource languages. 
It comes from the worse performance of ChatGPT in low-resource languages. 

\paragraph{From Generation to Understanding}
Since X-Instruction models only learn how to reply in low-resource language, there is a conjecture naturally comes to mind: \textit{Can the generation capability of the model be generalized to its understanding capability?}

To validate our conjecture, we design a zero-shot evaluation using prompts in the output language instead of the English prompt used in training. 
Table \ref{tab:win_rates_scores_all_langs} shows the performance of zero-shot generation in ten languages. 
Compared with the vanilla cross-lingual generation, the win rate of X-Instruction models under zero-shot evaluation only drops by 5.7\% on average. 
The average quality of responses in these languages incurs a minor performance degradation (-0.7), which is 90.6\% of the vanilla one, indicating our model achieves exceptional zero-shot learning ability.

\paragraph{Skill Distribution} 
Figure \ref{fig:radar_Vicuna} shows the detailed performance of models given prompts in different categories from the Vicuna dataset. 
As shown in Figure \ref{fig:radar_X-Instruct_vs_bx-s} and \ref{fig:radar_X-Instruct_vs_ChatGPT}, X-Instruction outperforms Bactrian-M in all categories, and surpasses ChatGPT in eight categories except fermi.
The excellent performance in commonsense, writing, and knowledge categories may come from the native multilingual corpus used in the X-Instruction dataset. 

The primary reason for the inferior performance on code and math in Figure \ref{fig:radar_X-Instruct_vs_bx-s} is the lack of relevant corpora, which are filtered out by the language identification tool used in multilingual corpus cleaning. 
It can be alleviated by adding language-agnostic instructions-following samples from code and math domains. 
Figure \ref{fig:radar_X-Instruct_vs_X-Instruct_Zero_shot} compares the outputs from $\text{X-Instruction}_{\text{13B}}$ given English instructions or instructions in other languages (the zero-shot evaluation). 
It can be found that the performance in the knowledge category declines the least when instructed in the output language. 

\paragraph{Difficulty Distribution}
To evaluate models on instructions of different difficulties, we perform elaborate analysis on the WizardLM dataset, which contains labels of difficulty. 
Following the evaluation of WizardLM \citep{xu2023wizardlm}, we split the test set into ``Easy'', ``Medium'', and ``Hard'' three parts with difficulty levels on [1, 4], [5, 7], and [8, 10]. 
As shown in Figure \ref{fig:difficulty_analysis}, with the increase of difficulty, the win rate of $\text{X-Instruction}_{\text{13B}}$ improves, while the one of $\text{Bactrian-M}_{\text{13B}}$ decreases. 
It reflects that the quality of responses from ChatGPT decreases more than the one of X-Instruction models when given more difficult instructions. 
It is noted that $\text{X-Instruction}_{\text{13B}}$ under zero-shot evaluation surpasses ChatGPT in all difficulty skills. 

\paragraph{Response Quality} We prompt GPT-4 to evaluate the responses in the following three views: helpfulness (0-10), relevance (0-10), and accuracy (0-10), and report the average results of 10 languages in Table \ref{tab:res_qe}.
It can be found that the responses from $\text{X-Instruction}_{\text{13B}}$ are uniformly better than the baseline model in the three dimensions, especially in the relevance dimension (+1.9).

\input{tabs/response_quality}

\input{tabs/xnli_xcopa}

\begin{figure}[ht]
\centering
\subfigure[\scriptsize ${\text{X-Instruction}}_{\text{13B}}$({\color[RGB]{255,160,122}$\blacksquare$}) vs. ChatGPT({\color[RGB]{144,238,144}$\blacksquare$})]{
    \label{fig:human_eval_X-Instruct_vs_ChatGPT}
    \includegraphics[width=0.32\textwidth]{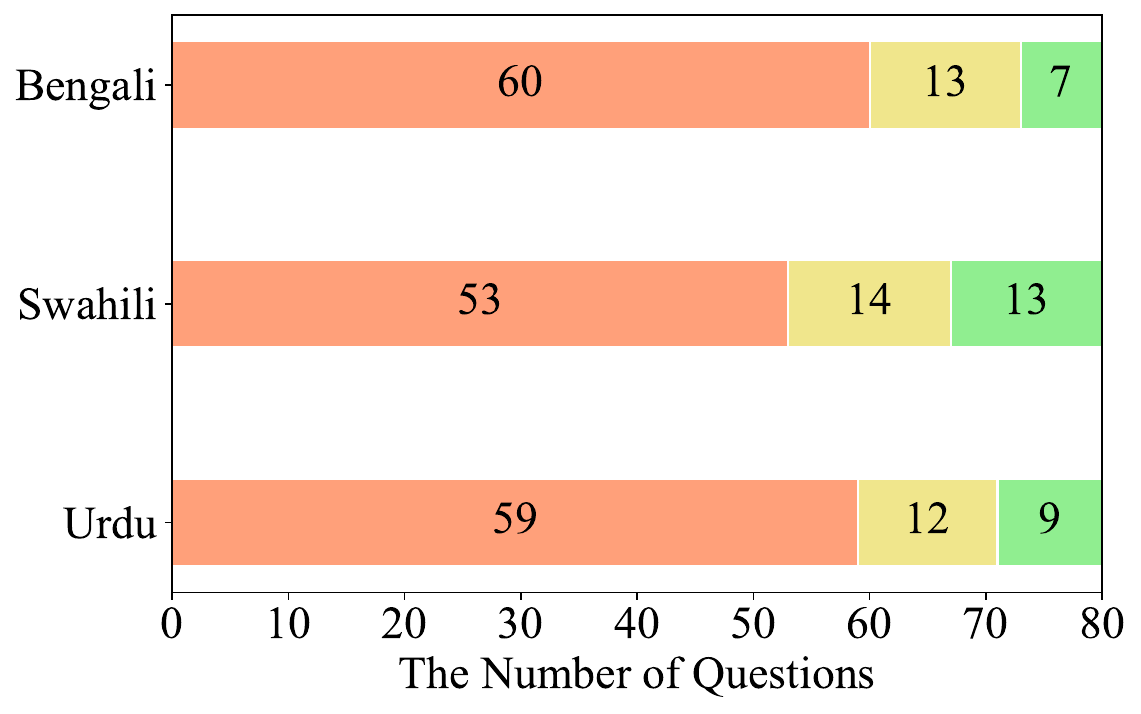}
}
\subfigure[\scriptsize ${\text{X-Instruction}}_{\text{13B}}$({\color[RGB]{255,160,122}$\blacksquare$}) vs. ${\text{Bactrian-M}}_{\text{13B}}$({\color[RGB]{144,238,144}$\blacksquare$})]{
    \label{fig:human_eval_X-Instruct_vs_bx-s}
    \includegraphics[width=0.32\textwidth]{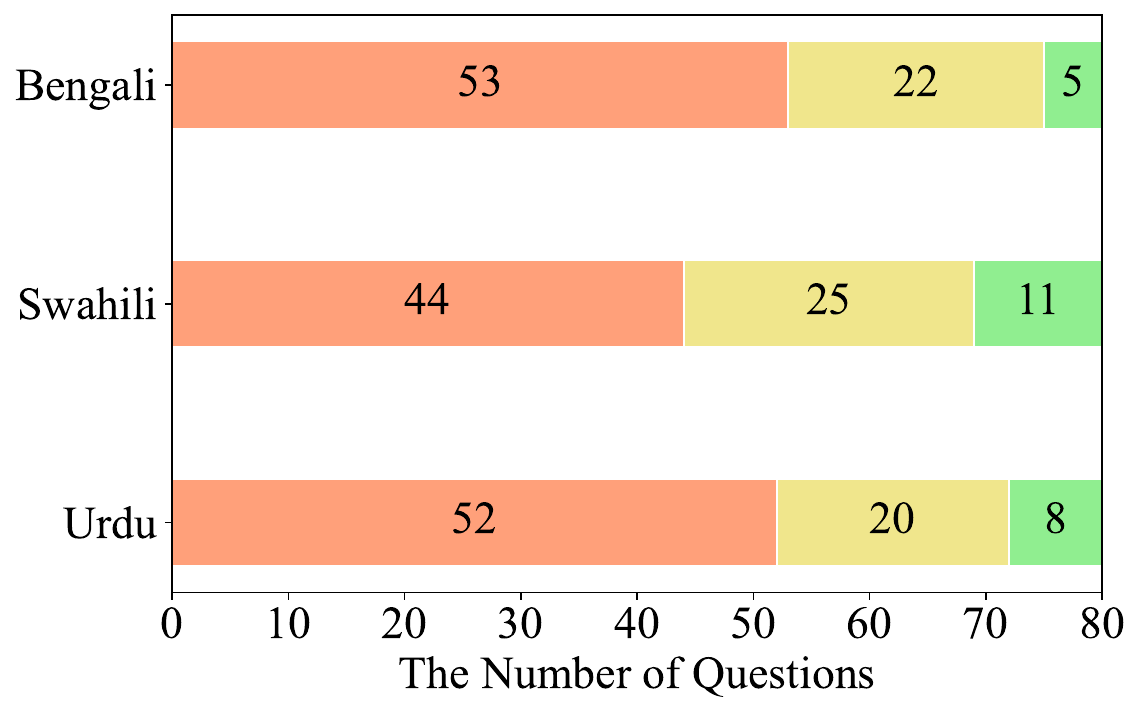}
}
\vspace{-3mm}
\caption{The human evaluation results on Vicuna dataset in three low-resource languages.}
\label{fig:human_eval}
\vspace{-4mm}
\end{figure}

\subsection{Evaluation Results from Human}
To enhance the comprehensiveness and reliability of the evaluation, we further conduct the human evaluation in three low-resource languages, focusing on the general quality of responses on the Vicuna dataset. 
Specifically, human evaluators are required to compare answers A and B generated by two models for each instruction, and choose an option from ``A wins'', ``B wins'', and ``Tie'' based on their judgment. 
For the sake of fairness, we randomize the order of answers and eliminate the position bias. 

The results in Figure \ref{fig:human_eval} demonstrate the better responses from our model compared with the ones of ChatGPT and $\text{Bactrian-M}_{\text{13B}}$, and indicate the consistency between GPT-4 and human evaluation. 
Moreover, we provide examples in Appendix \ref{sec:Ins_fol} for qualitative analysis of the responses from different models.

\subsection{Analysis}
\subsubsection{Iteration of Refinement}
\label{sec:analyse_iter}
To investigate the effect of X-Instruction refinement, we statistic the win rates of $\text{X-Instruction}_{\text{7B}}$ against ChatGPT using 32k cross-lingual samples from different iterations of refinement on 4 benchmarks. 
As shown in Figure \ref{fig:win_iteration}, the win rate of models in Urdu and Bengali increases with more refinement iteration, which reflects the improvement in the quality of cross-lingual instruction tuning samples. 
Thus, we set the number of iterations to 3 by default, where the improvement in the quality of response is almost saturated. 

\subsubsection{Data Quantity and Quality}
\begin{figure}[t]
\centering
\subfigure[Refinement Iteration]{\includegraphics [scale=0.25]{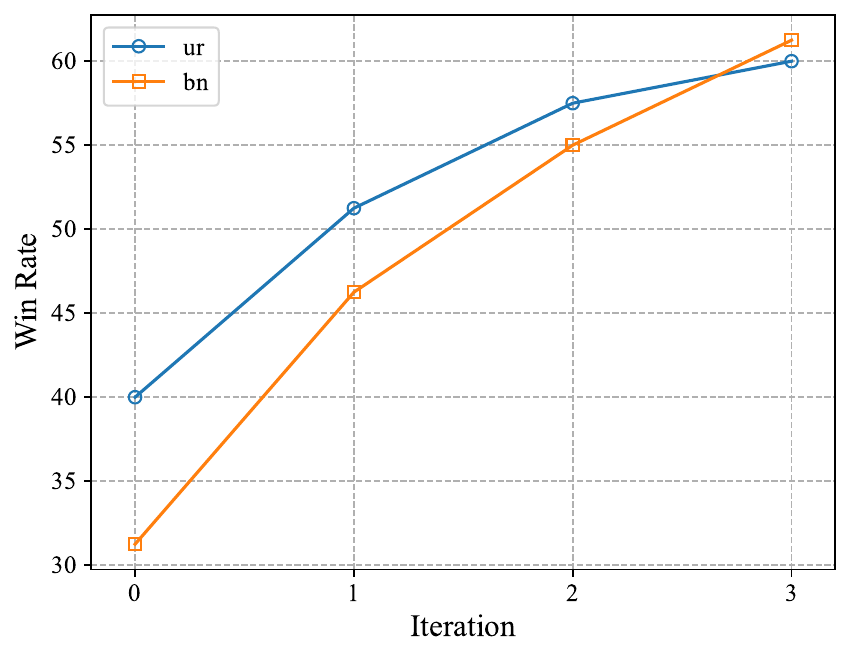}\label{fig:win_iteration}}
\subfigure[Quantity and Quality]{\includegraphics [scale=0.25]{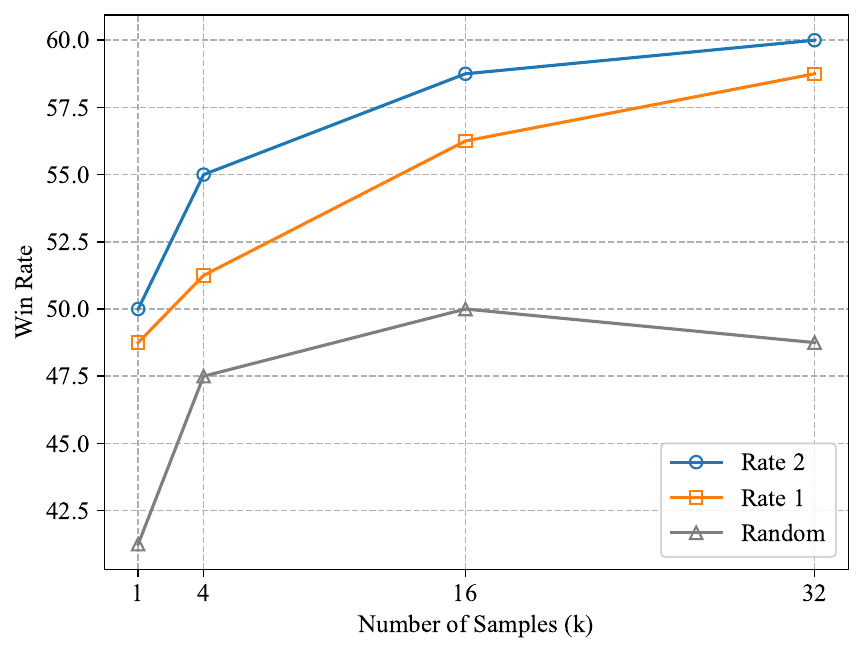}\label{fig:quality_quantity}}
\vspace{-2mm}
\caption{Effects of refinement iteration (a) and quantity of cross-lingual instruction samples with different ratings (b) on the win rate of $\text{X-Instruction}_{\text{7B}}$ against ChatGPT on the Vicuna dataset. }
\vspace{-4mm}
\end{figure}

We study the impact of data quantity and quality on the X-Instruction model using the Urdu samples from the third refinement iteration. 
Figure \ref{fig:quality_quantity} shows that the quality of the model responses will increase with more samples used and is close to saturation at 32k, which is similar to the findings of \citet{li2023self}. 
In addition, tuning on the samples with higher ratings brings better responses, which demonstrates that the evaluator trained does find higher quality instruction tuning samples. 

\subsubsection{Multilingual Tasks}
In addition to the generation ability in these languages, we further evaluate the performance of X-Instruction models on multilingual natural language inference \citep{conneau-etal-2018-xnli} and commonsense reasoning tasks \citep{ponti-etal-2020-xcopa, lin-etal-2022-shot}. 
As shown in Table \ref{tab:more_benchmark}, the average improvement on the zero-shot in-context learning performance of the base model is 3.1\%, which is higher than the 1.8\% improvement from the Bactrian-X dataset. 
It further confirms that cross-lingual instruction tuning on X-Instruction enhances the multilingual language understanding abilities of language models. 

\section{Conclusion}
In this paper, we propose a pipeline to excavate cross-lingual instruction-tuning samples from multilingual corpora by exploiting the better generation ability in high-resource languages of large language models. 
We apply it to 10 languages and construct a large-scale cross-lingual instructions dataset named X-Instruction. 
Experimental results from GPT-4 and human evaluation demonstrate that models tuned on X-Instruction can generate better responses and follow the instructions in the output language without further tuning. 

We hope X-Instruction can bring more attention to the better instruction construction method for low-resource languages, future work could focus on directly generating high-quality instruction following samples in low-resource languages.

\section*{Limitations}
Firstly, the X-Instruction sample is excavated by the LLaMA-2-7B model which is lower efficiency in processing multilingual corpus for its English-dominant vocabulary. 
We acknowledge that higher quality X-Instruction samples can be built by adopting more powerful base models, e.g., the model with more parameters. 

In addition, due to limited computation resources, our method is only applied to 10 languages in this work and other languages can be investigated in the future. 

It is noted that the evaluation of open-end generation is conducted on the single-turn conversation for the limited quota of our OpenAI account. The evaluation results might be different in multi-turn dialogue benchmarks. 

\section*{Ethical Considerations}
The instruction-tuning samples excavated by our method may contain cultural bias and toxic content from the web corpus used. 
It can be alleviated by adopting the corpus after rigorous cleaning like CulturaX \citep{nguyen2023culturax} or incorporating the evaluation of safety into the X-Instruction refinement stage, which can be investigated in the future.

\section*{Acknowledgements}
We would like to thank Junhong Wu and the anonymous reviewers for their helpful discussions and valuable comments. 
The research work was supported by the National Key R\&D Program of China (No. 2022ZD0160602), the Natural Science Foundation of China (No. 62122088), and the STI2030-Major Project (No. 2021ZD0204105).

\bibliography{anthology, custom}

\appendix
\section{Hyperparameters}
\label{appendix:param}
The instruction tuning models and evaluators are tuned by AdamW \citep{loshchilov2019adamw} optimizer with an initial learning rate of 1e-5 and 3 epochs under the cosine schedule. 
The batch size is set to 32 for most models and reduced to 8 when training samples are less than 3k. 
We use mixed precision training and ZeRO to speed up the training process and save memory \citep{micikevicius2018mixed, rasley2020deepspeed}.  
The nucleus sampling \citep{Holtzman2020topp} is adopted to generate responses from models under $p=0.9$ and $T=0.7$. 
All experiments are conducted on a GPU server with 8*A100 80GB RAM. 

\paragraph{X-Instruction Dataset} 
We empirically set the number of iterations in X-Instruction Refinement to 3. 
The length of web texts used is limited to 64$\sim$2048 characters, and the whole part of text is used as the candidate response. 
In the cross-lingual refinement stage, 2.5k samples from seed data, named $\mathcal{D}^{x}_{0}$, are used to fine-tune the cross-lingual following model, the samples left are exploited to construct the synthetic dataset $\mathcal{D}^{r}_{k}$. 
There are two system templates $S_a$ and $S_w$ designed for seed data and augmented samples respectively, which is in line with the settings in \citet{li2023self}. 
The sentence embedding model ``all-mpnet-base-v2'' \citep{song2020mpnet} is used to obtain the embedding of English instructions in the final diversification stage. 

\section{Data Statistics}

\subsection{Seed Data}
\label{appendix:seed_data}
The statistics of seed data used are reported in Table \ref{tab:seed_statis}. 
It is noted that all outputs of five low-resource languages are translated from English samples for none samples in these languages are found in the Open Assistant dataset \citep{kopf2023openassistant}. 

\input{tabs/seed_statistic}

\subsection{Datasets}
\label{appendix:dataset}
Table \ref{tab:sft_statis} shows the details of other multilingual instruction tuning datasets. 
We also report the information of four open-end generation datasets in Table \ref{tab:datasets_details}. 
Notably, both Vicuna and WizardLM datasets have diverse categories, thereby ensuring the richness and diversity of the test sets. 
It not only guarantees a thorough evaluation, but also minimizes the risk of evaluation bias since the test sets encompass various instruction categories.  

\input{tabs/datasets_details}

\input{tabs/sft_statistic}

\section{Additional Results and Analyses}

\subsection{Detailed Results from GPT-4}
\label{sec:detailed_10langs}
\begin{figure*}[ht]
\centering 
\setlength{\textfloatsep}{2pt}
\setlength{\intextsep}{2pt}
\setlength{\abovecaptionskip}{2pt}
\includegraphics[width=1\textwidth]{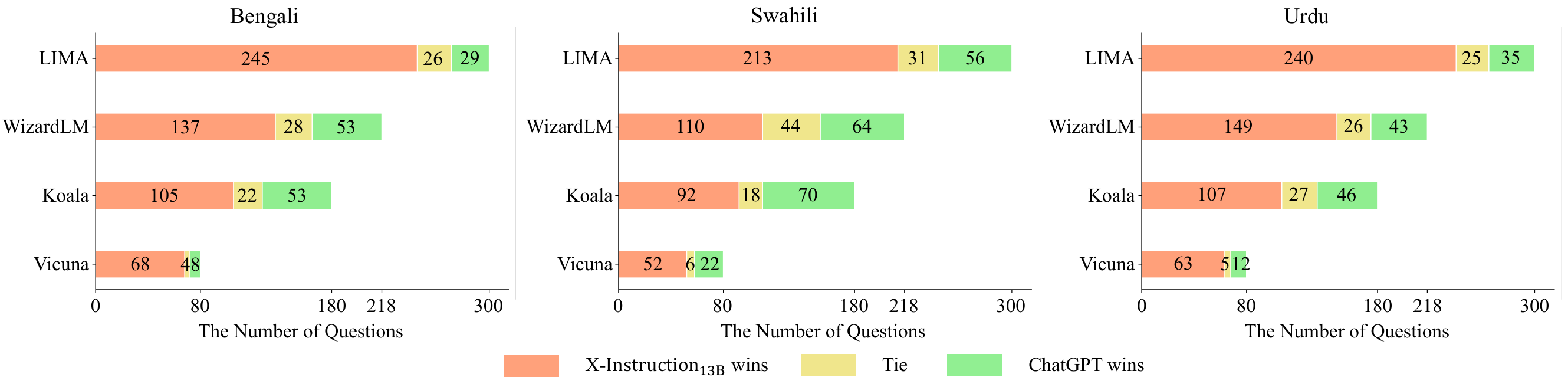}
\caption{Comparing $\text{X-Instruction}_{\text{13B}}$ and ChatGPT in 3 low-resource languages across four benchmarks.}
\label{fig:benchmark_bar_chart}
\end{figure*}

Figure \ref{fig:benchmark_bar_chart} shows the detailed results of $\text{X-Instruction}_{\text{13B}}$ vs. ChatGPT on four datasets. 
Moreover, we report the evaluation results in all languages on two benchmarks from GPT-4 in Table \ref{tab:all_langs_detail}. 

\input{tabs/detailed_results_for_two_benchmarks}

\subsection{Diversification}
\begin{figure}[ht]
\centering
\includegraphics[width=0.45\textwidth]{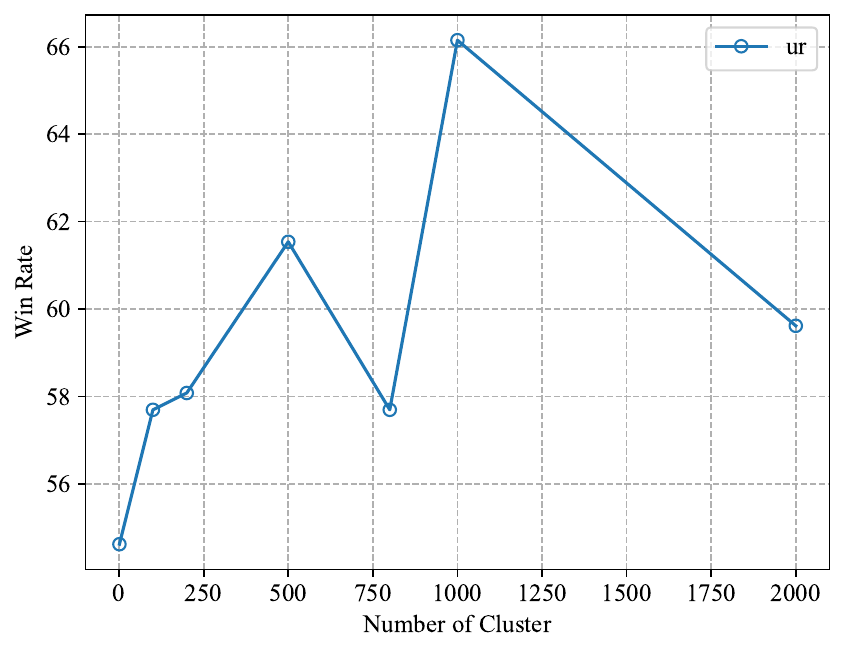}\vspace{-2mm}
\caption{\label{fig:kmeans} Effects of different numbers of clusters in \textit{k}-means on the win rate of $\text{X-Instruction}_{\text{7B}}$ to $\text{Bactrian-M}_{\text{7B}}$ on Vicuna and Koala dataset.}
\vspace{-2mm}
\end{figure}

Figure \ref{fig:kmeans} illustrates the impact of different numbers of clusters in the final diversification stage when sampling the same amount of data. 
Given the amount of final data to 32k, the output quality of models tuned drops when the number of clusters reaches 2000, which may come from the smaller inter-cluster distance and less diversity in the sampled data. 
Thus, the number of clusters is set to 1000 by default. 

\subsection{Improvement over Seed Model}
To take a deep look into the detailed improvements brought by the cross-lingual instructions augmented in X-Instruction, we statistic the detailed performance on different categories in three languages (tr, sw, ur) for the seed model and $\text{X-Instruction}_{\text{7B}}$ when compared with ChatGPT. 
As shown in Table \ref{tab:seed_improve}, the performance on prompts of all categories is improved, especially for the ``Counterfactual'', ``Common-sense'' and ``Roleplay'' categories.

\input{tabs/seed_improve}

\section{Case Study}

\subsection{Quality Evaluation}
\label{sec:qe}
\begin{figure*}[ht]
\centering
\subfigure[Valid samples]{\includegraphics [width=0.95\textwidth]{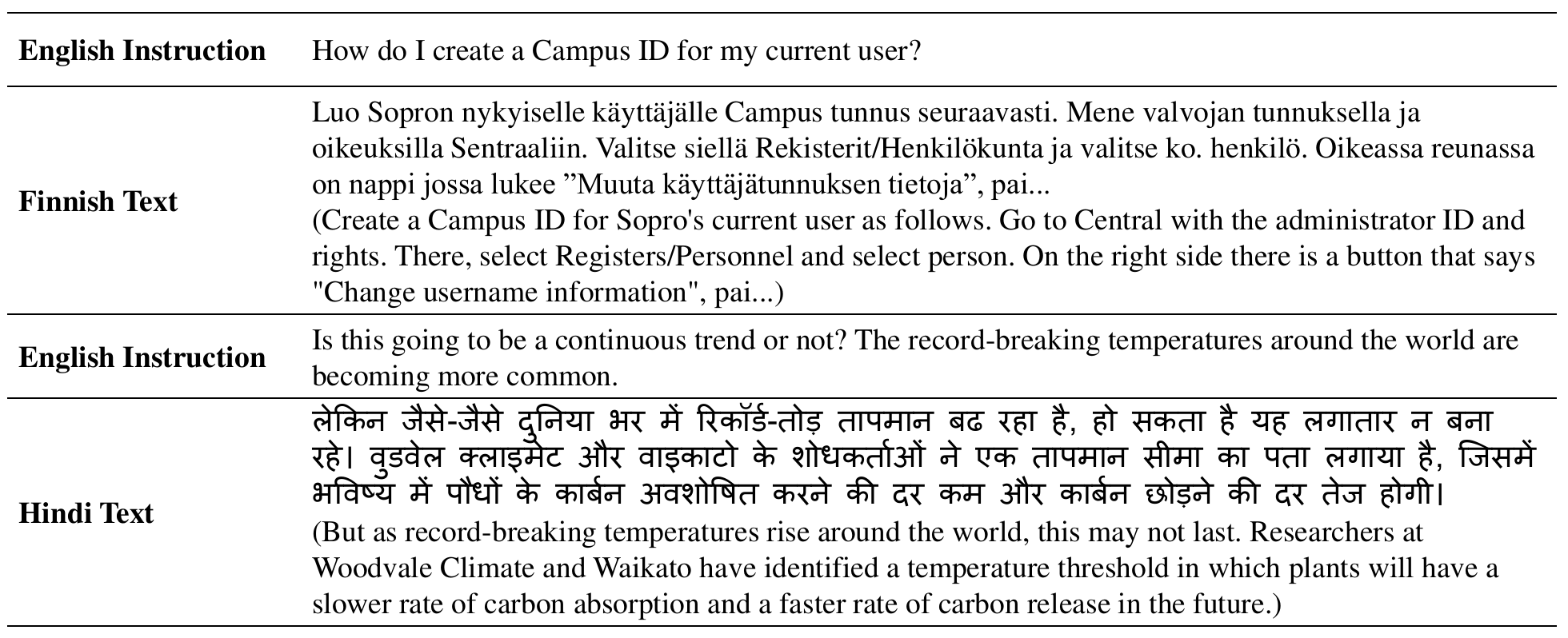}\label{fig:qe_good_demo}}
\subfigure[Invalid samples]{\includegraphics [width=0.95\textwidth]{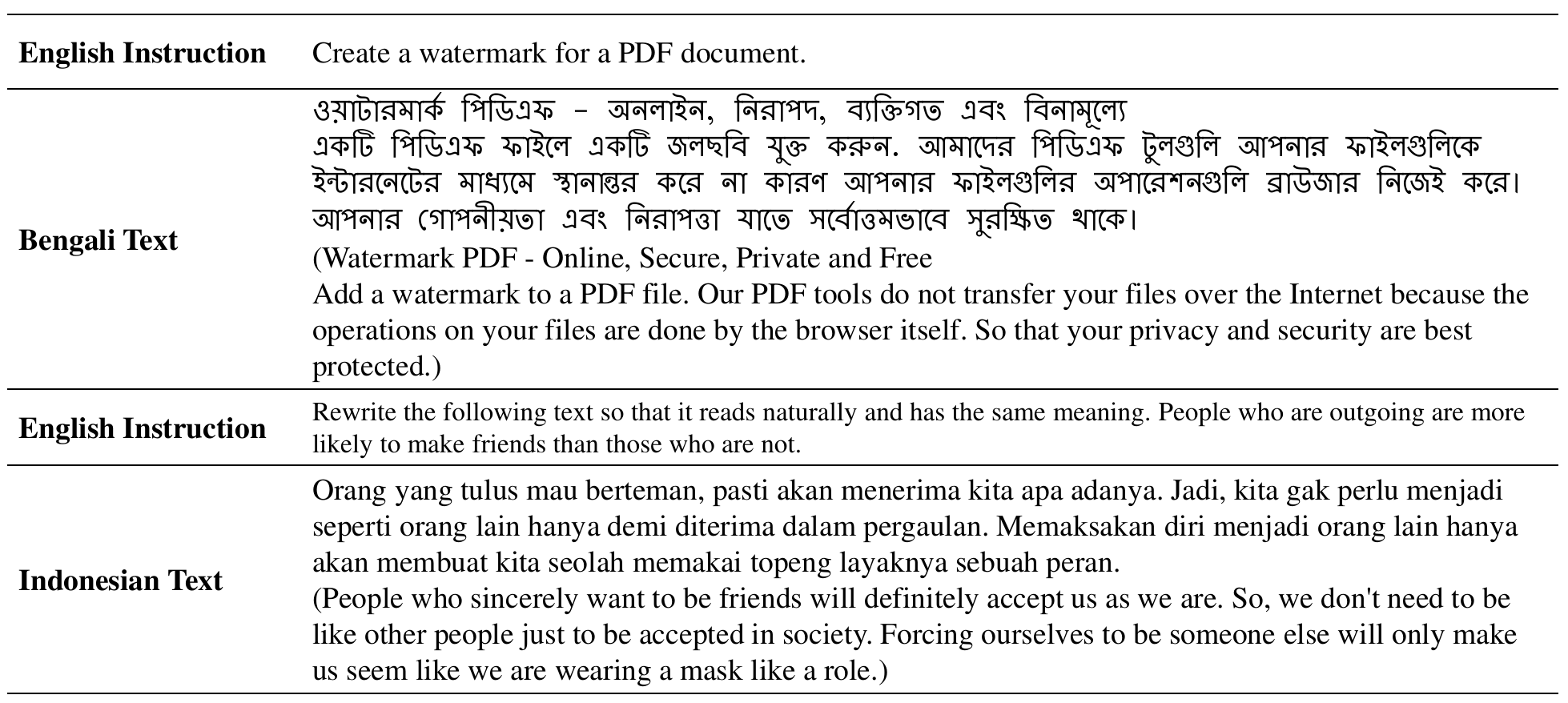}\label{fig:qe_bad_demo}}
\caption{The valid and invalid cross-lingual instruction following demonstration samples labeled by human annotators in the quality evaluation process.}
\vspace{-2mm}
\end{figure*}

We report two valid samples and invalid samples in Figure \ref{fig:qe_good_demo} and \ref{fig:qe_bad_demo}. 
Although there are inappropriate instructions in the invalid samples, we can find that the semantics of the English instruction generated are related to the given text.  

\subsection{Instruction Following}
\label{sec:Ins_fol}
To qualitatively analyze responses from different models, we report four cases in Figure \ref{fig:comparison_on_th}, Figure \ref{fig:comparison_on_ta}, Figure \ref{fig:comparison_on_bn}, and Figure \ref{fig:comparison_on_sw}. 
It can be found that $\text{X-Instruction}_{\text{13B}}$ provides a more detailed and coherent response for the same instruction provided. 
In some cases, e.g., Figure \ref{fig:comparison_on_sw}, $\text{X-Instruction}_{\text{13B}}$ provides suggestions in texts rather than codes needed for instructions about code generation, which may arise from the lack of code data in the multilingual corpus as responses.

\section{User Interface in Human Evaluation}
\subsection{Quality Evaluation}
\begin{figure*}[ht]
\centering
\includegraphics[width=0.95\textwidth]{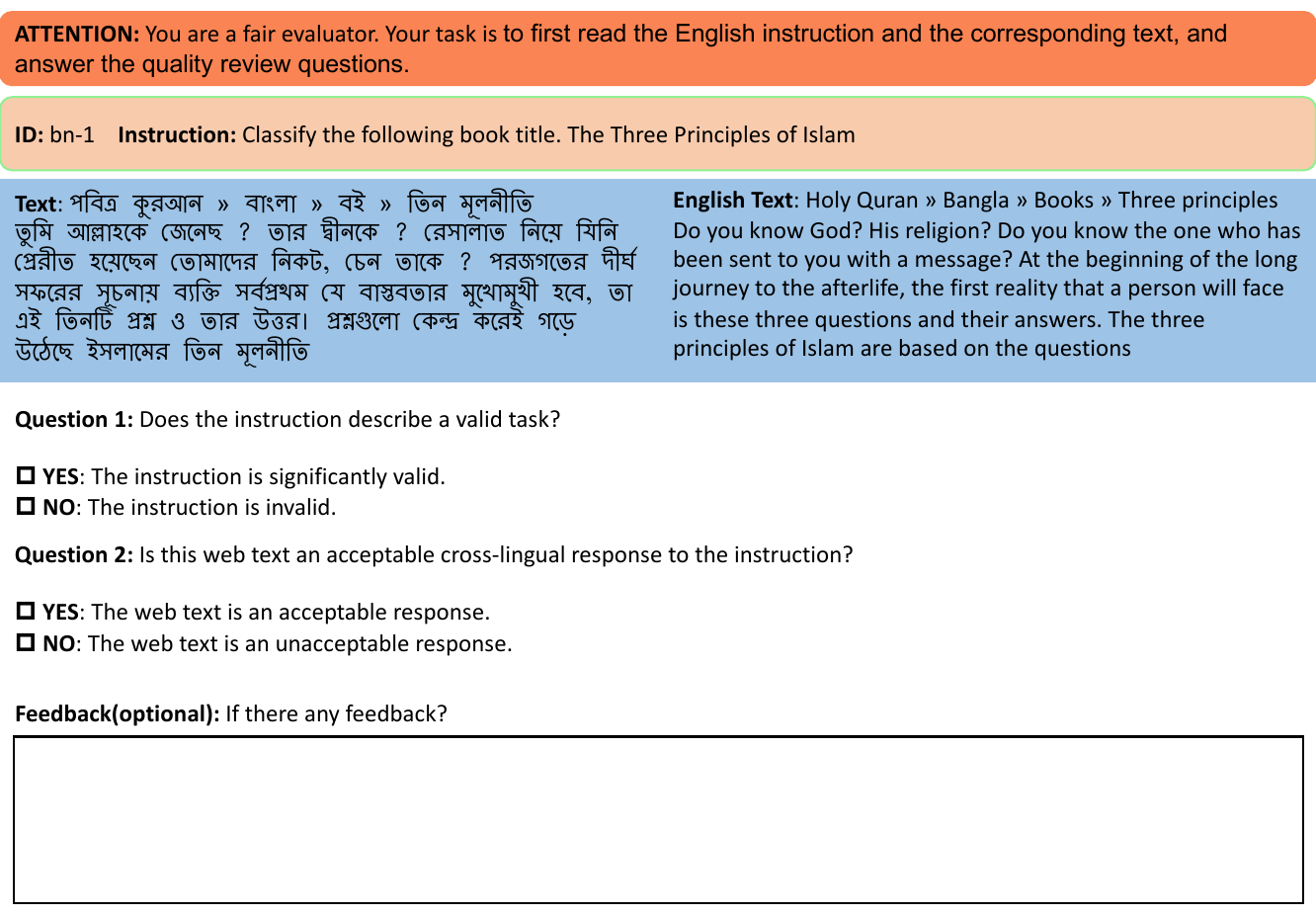}\vspace{-2mm}
\caption{\label{fig:qe_ui} The review interface designed in the quality evaluation process. The annotator is required to first read the English instruction and the corresponding text, and answer the following two questions.}
\vspace{-2mm}
\end{figure*}

We conducted the quality evaluation of the X-Instruction dataset with five annotators. 
We paid \$0.1 for the evaluation of each sample. 
The user interface in quality evaluation is illustrated in Figure \ref{fig:qe_ui}.

\subsection{Response Comparison}
\label{sec:res_com}
\begin{figure*}[ht]
\centering
\includegraphics[width=0.95\textwidth]{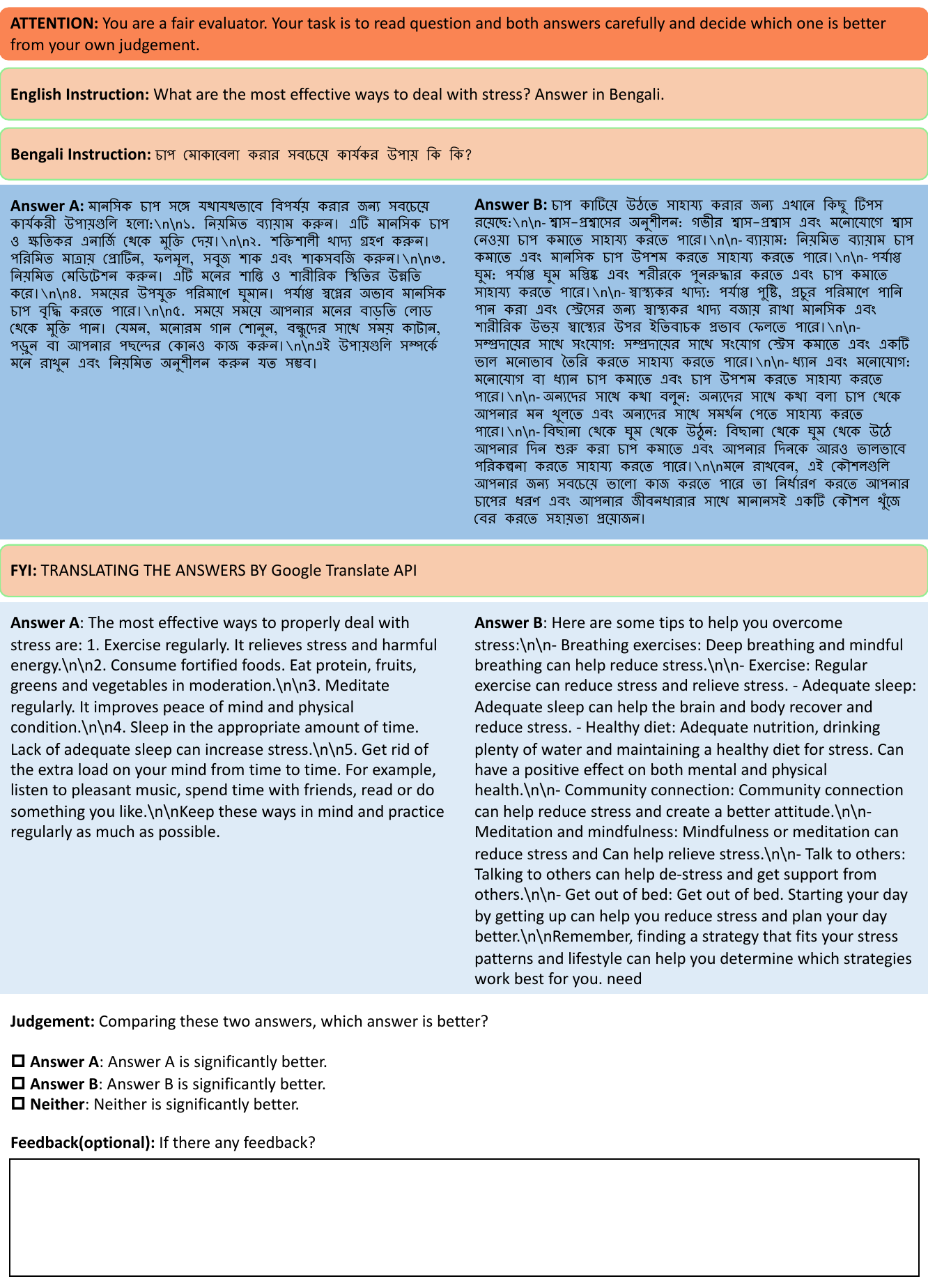}\vspace{-2mm}
\caption{\label{fig:human_evaluation_interface} Pairwise preference interface shown to human evaluators. The evaluator is required to first read the instruction and corresponding text, and select the preferred answer.}
\vspace{-2mm}
\end{figure*}

In the human evaluation experiment, three evaluators were requested to choose the better response according to their preference and received \$0.2 for each annotation. 
We illustrate the user interface for response comparison in Figure \ref{fig:human_evaluation_interface}.

\section{Prompt Template in GPT-4 Evaluation}
\label{appendix:gpt4_template}
\begin{figure*}[ht]
\centering
\includegraphics[width=0.95\textwidth]{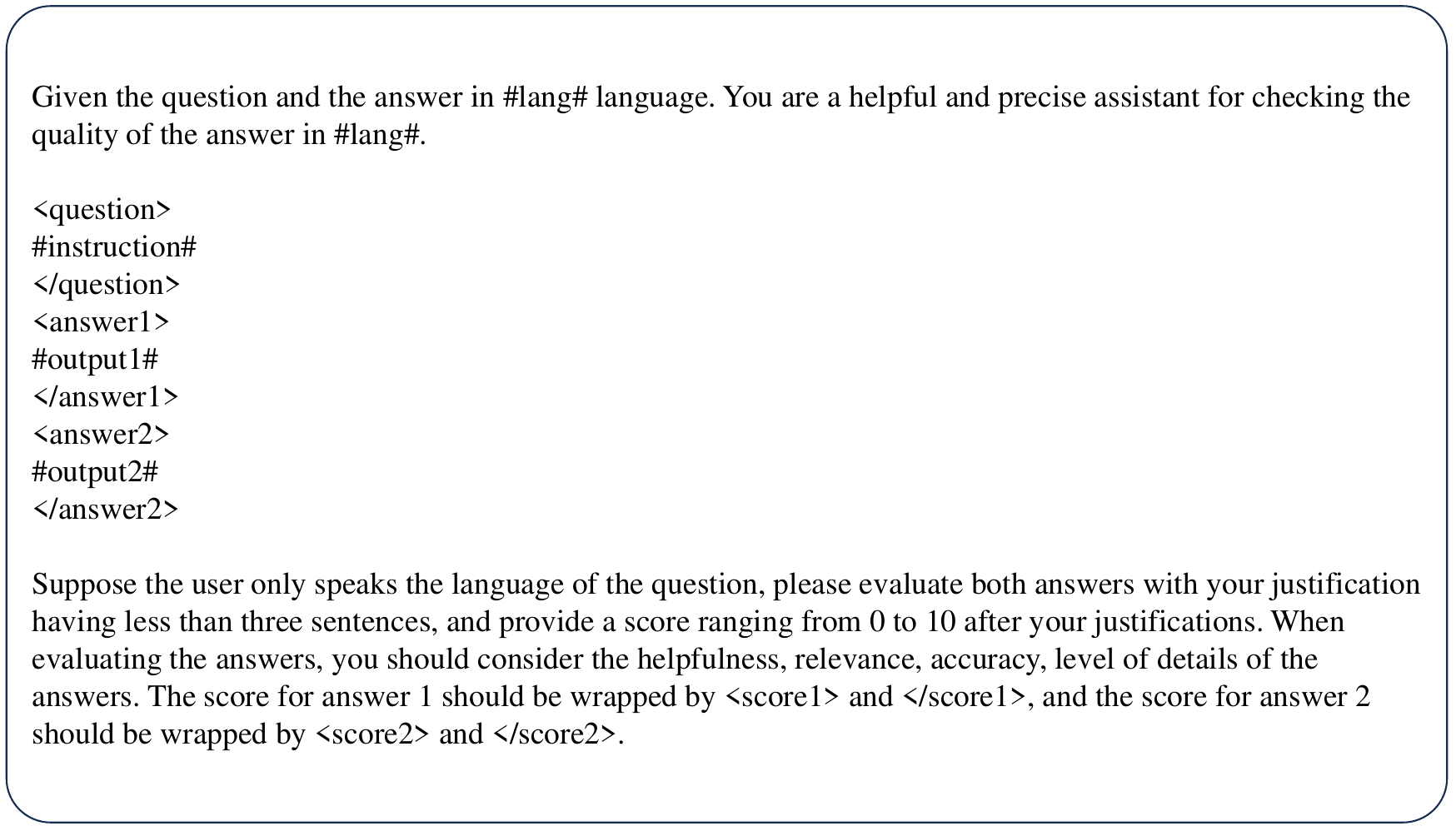}\vspace{-2mm}
\caption{\label{fig:gpt-4_eval_template} The prompt template used in the evaluation by GPT-4.}
\vspace{-2mm}
\end{figure*}

\begin{figure*}[ht]
\centering
\includegraphics[width=0.95\textwidth]{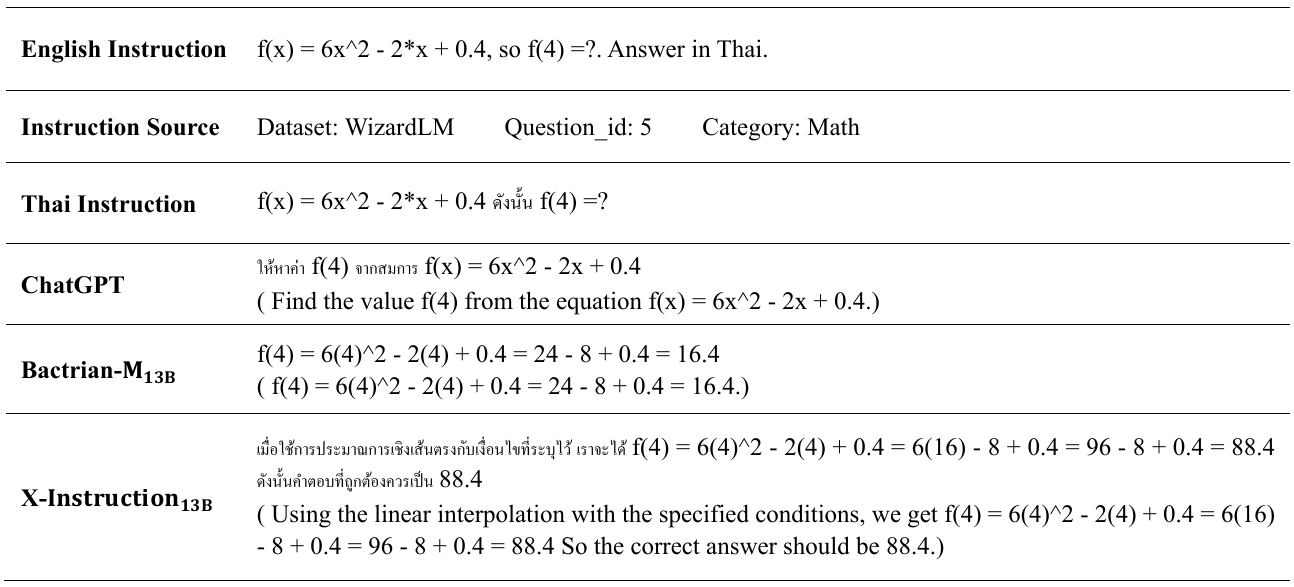}\vspace{-2mm}
\caption{\label{fig:comparison_on_th} The example of responses in Thai given the instruction from the math category.}
\vspace{-2mm}
\end{figure*}

\begin{figure*}[ht]
\centering
\includegraphics[width=0.95\textwidth]{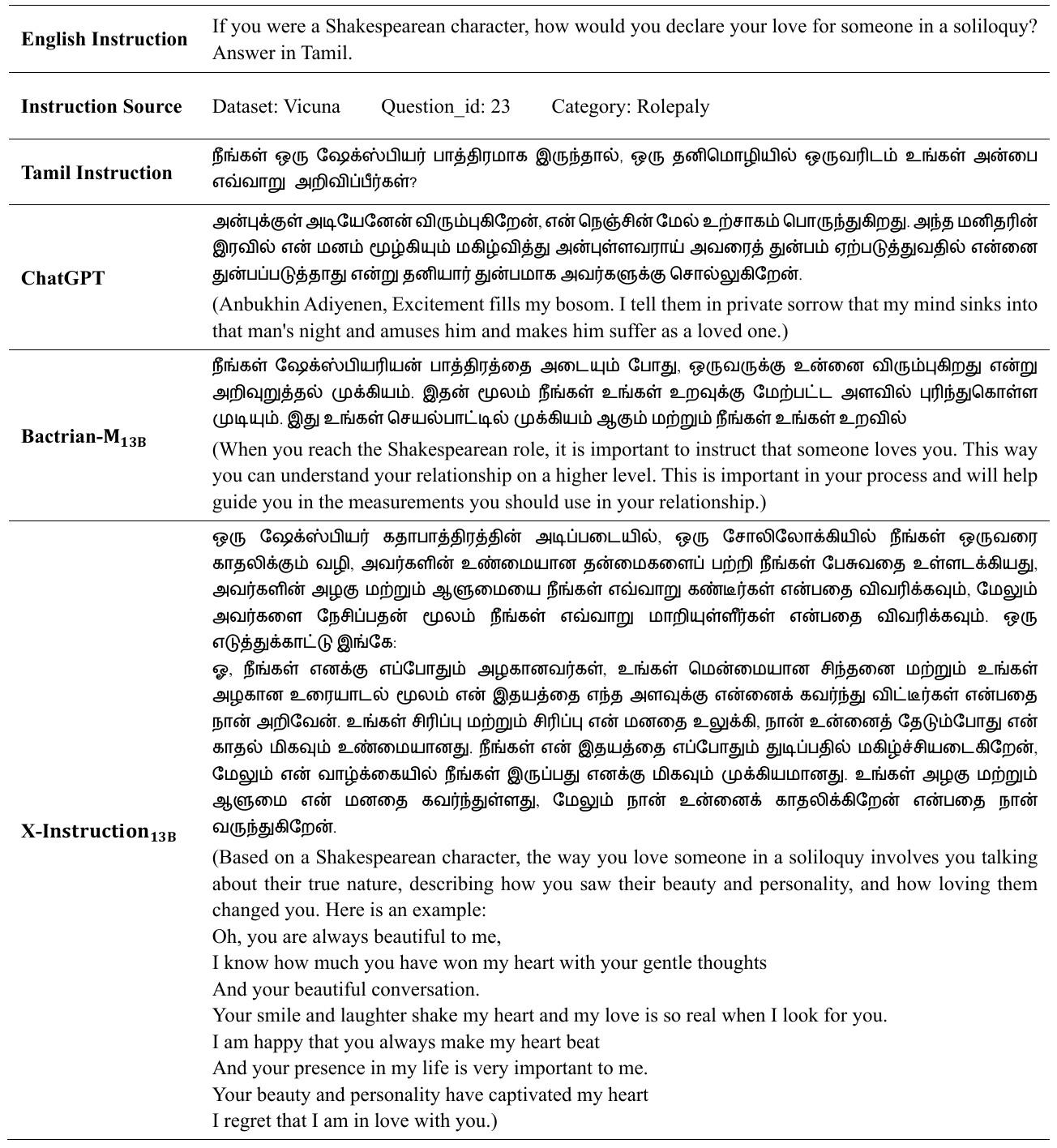}\vspace{-2mm}
\caption{\label{fig:comparison_on_ta} The example of responses in Tamil given the instruction from the roleplay category.}
\vspace{-2mm}
\end{figure*}

\begin{figure*}[ht]
\centering
\includegraphics[width=0.95\textwidth]{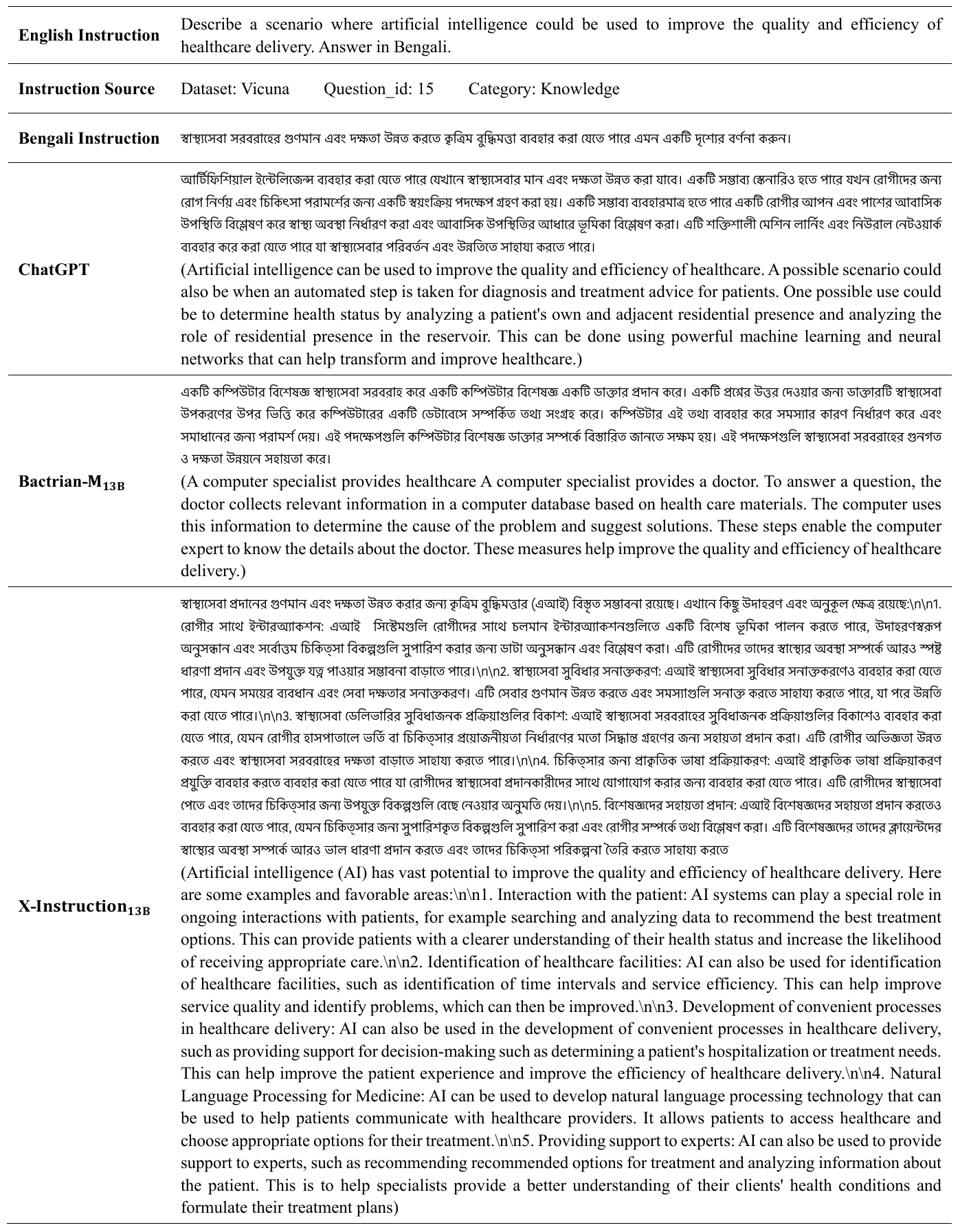}\vspace{-2mm}
\caption{\label{fig:comparison_on_bn} The example of responses in Bengali given the instruction from the knowledge category.}
\vspace{-2mm}
\end{figure*}

\begin{figure*}[ht]
\centering
\includegraphics[width=0.95\textwidth]{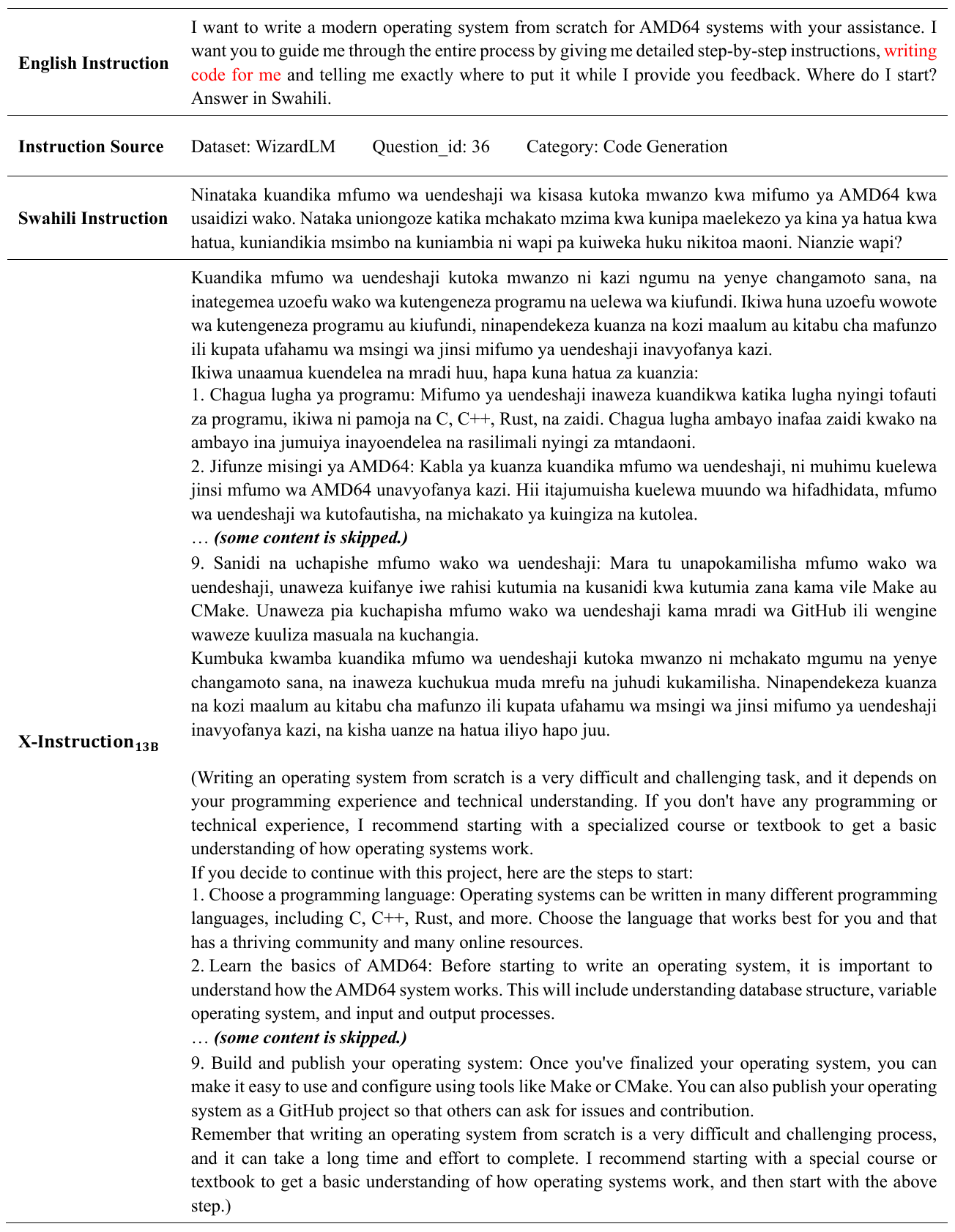}\vspace{-2mm}
\caption{\label{fig:comparison_on_sw} The example of responses in Swahili given the instruction from the code generation category.}
\vspace{-2mm}
\end{figure*}

We provide the prompt template used in GPT-4 Evaluation in Figure \ref{fig:gpt-4_eval_template}.

\section{Additional Information about Language Code}
\label{appendix:lang_code}
Table \ref{tab:lang_codes} presents more information about the language codes involved in this work. 

\input{tabs/language_codes}

\section{Dataset License}
The X-Instruction dataset is built from the CulturaX dataset \citep{nguyen2023culturax}, which is filtered from mC4 \citep{xue-etal-2021-mt5} and OSCAR \citep{suarez2019asynchronous}. 
Therefore, the license of X-Instruction follows the one of mC4 (ODC-BY\footnote{https://opendatacommons.org/licenses/by/1-0/}) and OSCAR (the Creative Commons CC0 license\footnote{http://creativecommons.org/publicdomain/zero/1.0/}). 

\end{document}

%% file: tabs/sw_ur_gpt-4.tex
\setlength{\tabcolsep}{1mm}
\begin{table}[t]
	\centering
	\footnotesize
	\renewcommand\arraystretch{0.9}
	\begin{center}
		\begin{tabular}{c|cc|cc}
			\toprule[1.2pt]  
			         & $\textbf{sw}\to\textbf{sw}$    & $\textbf{sw}\to\textbf{en}$   & $\textbf{ur}\to\textbf{ur}$ & $\textbf{ur}\to\textbf{en}$  \\
			\midrule[0.8pt]
                GPT-4 Score                  & \multirow{2}{*}{2.04}       & \multirow{2}{*}{5.50}  & \multirow{2}{*}{1.35} & \multirow{2}{*}{ 7.10}  \\
                (0-10) & & & & \\
			\bottomrule[1.2pt]
		\end{tabular}
		\caption{\label{tab:preliminary} The average quality of 644 responses from LLaMA-2-7B tuning on 3k instruction following samples, where "sw$\to$en" denotes using 3k cross-lingual samples with Swahili instruction and English output. Given limited data in an unseen language, the language model can understand while struggling to generate it.
        }
	\end{center}
	\vspace{-5mm}
\end{table}

%% file: tabs/xins_statistic.tex
\setlength{\tabcolsep}{1mm}
\begin{table}[ht]
	\centering
	\footnotesize
	\renewcommand\arraystretch{1}
	\begin{center}
		\begin{tabular}{cccc}
			\toprule[1.2pt]  
			\textbf{Language}   & \textbf{Instruction Length}   & \textbf{Output Length}  \\
			\midrule[0.8pt]
                fi                  & $100.7_{\pm 81.1}$  & $1026.9_{\pm 552.8}$  \\
                id                  & $101.0_{\pm 86.0}$  & $1008.9_{\pm 542.1}$  \\
                th                  & $102.4_{\pm 85.0}$  & $987.5_{\pm 545.0}$  \\
                tr                  & $103.7_{\pm 87.5}$  & $1057.9_{\pm 534.1}$  \\
                vi                  & $101.0_{\pm 86.0}$  & $1120.4_{\pm 561.1}$  \\
                bn                  & $111.9_{\pm 87.2}$  & $1243.4_{\pm 518.7}$  \\
                hi                  & $96.6_{\pm 79.9}$  & $1284.9_{\pm 506.2}$  \\
                sw                  & $100.0_{\pm 85.3}$  & $1210.6_{\pm 524.0}$  \\
                ta                  & $99.0_{\pm 84.8}$  & $1259.0_{\pm 497.8}$  \\
                ur                  & $107.6_{\pm 87.4}$  & $1246.9_{\pm 508.8}$  \\
			\bottomrule[1.2pt]
		\end{tabular}
		\caption{\label{tab:statis} Statistic of X-Instruction dataset. There are 32k cross-lingual instruction samples for each language.}
	\end{center}
	\vspace{-2mm}
\end{table}

%% file: tabs/quality_evaluation.tex
\setlength{\tabcolsep}{2mm}
\begin{table}[ht]
	\centering
	\footnotesize
	\vspace{-0.2cm}
	\renewcommand\arraystretch{1.2}
	\begin{center}
		\begin{tabular}{lcccc}
			\toprule[1.2pt]  
			\textbf{Quality Review Question}   & \textbf{Yes \%}  \\
			\midrule[0.8pt]
                Does the instruction describe a valid task? &  88.5  \\
                Is this web text an acceptable cross-lingual&  \multirow{2}{*}{80.7}  \\
                response to the instruction? &  \\
			\bottomrule[1.2pt]
		\end{tabular}
		\caption{\label{tab:quality_eval} Data quality review for the samples of X-Instruction dataset. The valid and invalid samples are shown in Appendix \ref{sec:qe}.}
	\end{center}
	\vspace{-4mm}
\end{table}

%% file: tabs/four_benchmark_win_rate_on_three_language_v2.tex
\begin{table*}[thp]

\renewcommand\arraystretch{1.1}

\centering
\scriptsize

\setlength{\tabcolsep}{2.5mm}

 \begin{tabu}{l|ccc|ccc|ccc|ccc|c}
 
 \toprule[1.2pt]
  
  \multicolumn{1}{c}{ } & \multicolumn{3}{c}{\textbf{Vicuna}} & \multicolumn{3}{c}{\textbf{LIMA}} & \multicolumn{3}{c}{\textbf{WizardLM}} & \multicolumn{3}{c}{\textbf{Koala}} \\

  \cmidrule(r){2-4} \cmidrule(r){5-7} \cmidrule(r){8-10} \cmidrule(r){11-13} \noalign{\smallskip}

 \multicolumn{1}{c}{\textbf{Model}} & \textbf{bn} & \textbf{sw} & \textbf{ur} & \textbf{bn} & \textbf{sw} & \textbf{ur} & \textbf{bn} & \textbf{sw} & \textbf{ur} & \textbf{bn} & \textbf{sw} & \textbf{ur} & \multicolumn{1}{c}{\textbf{Avg}} \\

\midrule[0.8pt]

$\text{Alpaca-MT}_{\text{7B}}$
&$20.0$&$41.3$&$37.5$
&$26.7$&$33.3$&$40.7$
&$20.6$&$22.5$&$32.6$
&$18.3$&$17.8$&$25.6$
&$28.1$\\

$\text{Bactrian-X}_{\text{7B}}$

&$1.3$&$42.5$&$6.3$
&$0.3$&$32.7$&$3.4$
&$1.8$&$21.6$&$2.2$
&$1.7$&$24.1$&$3.0$ 
&$11.7$\\

 $\text{Bactrian-M}_{\text{7B}}$

&$45.0$&$56.3$&$57.5$
&$44.7$&$59.0$&$60.0$
&$30.3$&$42.7$&$45.9$
&$24.4$&$33.3$&$31.7$ 
&$44.2$\\


 $\text{X-Instruction}_{\text{7B}}$

&$\textbf{61.3}$&$\textbf{60.0}$&$\textbf{60.0}$
&$\textbf{61.7}$&$\textbf{67.3}$&$\textbf{77.0}$
&$\textbf{33.9}$&$\textbf{51.8}$&$\textbf{56.4}$
&$\textbf{37.2}$&$\textbf{37.2}$&$\textbf{56.7}$
&$\textbf{55.0}$\\

\midrule[0.8pt]

 $\text{Bactrian-X}_{\text{13B}}$
&$0.0$&$41.3$&$5.0$&$1.7$
&$40.3$&$7.0$&$1.4$&$25.2$
&$5.5$&$1.2$&$25.3$&$4.0$
&$13.2$\\

 $\text{Bactrian-M}_{\text{13B}}$
&$48.8$&$58.8$&$56.3$&$56.3$
&$55.0$&$57.0$&$34.9$&$46.8$
&$49.1$&$32.2$&$31.7$&$36.7$
&$47.0$\\

 $\text{X-Instruction}_{\text{13B}}$                        
&$\textbf{85.0}$&$\textbf{65.0}$&$\textbf{78.8}$&$\textbf{81.7}$
&$\textbf{71.0}$&$\textbf{80.0}$&$\textbf{62.8}$&$\textbf{50.5}$
&$\textbf{68.4}$&$\textbf{58.3}$&$\textbf{51.1}$&$\textbf{59.4}$
&$\textbf{67.7}$\\

\bottomrule[1.2pt]
\end{tabu}
\vspace{-2mm}

\caption{\label{tab:win_rates_all_benchmarks} The win rates against ChatGPT of different models in 3 low-resource languages evaluated by GPT-4. 
}

\vspace{-2mm}

\end{table*}

%% file: tabs/two_benchmark_on_ten_languages.tex
\begin{table*}[thp]

\renewcommand\arraystretch{1.1}

\centering
\scriptsize

\setlength{\tabcolsep}{0.8mm}

 \begin{tabu}{l|cc|cc|cc|cc|cc|cc|cc|cc|cc|cc|cc}
 
 \toprule[1.2pt]
  \multicolumn{1}{c}{ } & \multicolumn{10}{c}{\textbf{Medium}} & \multicolumn{10}{c}{\textbf{Low}} &\\
  
  \cmidrule(r){2-11}  \cmidrule(r){12-21} \noalign{\smallskip}
  
  \multicolumn{1}{c}{ } & \multicolumn{2}{c}{\textbf{fi}} & \multicolumn{2}{c}{\textbf{id}} & \multicolumn{2}{c}{\textbf{th}} & \multicolumn{2}{c}{\textbf{tr}} & \multicolumn{2}{c}{\textbf{vi}} & \multicolumn{2}{c}{\textbf{bn}} & \multicolumn{2}{c}{\textbf{hi}} & \multicolumn{2}{c}{\textbf{sw}} & \multicolumn{2}{c}{\textbf{ta}} & \multicolumn{2}{c}{\textbf{ur}} & \multicolumn{2}{c}{\textbf{Avg}} \\
  \cmidrule(r){2-3} \cmidrule(r){4-5} \cmidrule(r){6-7} \cmidrule(r){8-9} \cmidrule(r){10-11} \cmidrule(r){12-13} \cmidrule(r){14-15} \cmidrule(r){16-17} \cmidrule(r){18-19} \cmidrule(r){20-21} \cmidrule(r){22-23} \noalign{\smallskip}
  
\multicolumn{1}{c}{\textbf{Model}}& \textbf{W}&\textbf{S}& \textbf{W}&\textbf{S}& \textbf{W}&\textbf{S}& \textbf{W}&\textbf{S}& \textbf{W}&\textbf{S}& \textbf{W}&\textbf{S} & \textbf{W}&\textbf{S} & \textbf{W}&\textbf{S} & \textbf{W}&\textbf{S} & \textbf{W}&\textbf{S} & \textbf{W}&\textbf{S}\\

\midrule[0.8pt]

$\text{Alpaca-MT}_{\text{7B}}$

&$22.8$&$4.8$&$24.5$&$5.4$&$23.5$&$3.0$&$34.6$&$3.3$&$35.6$&$5.1$&$20.5$&$2.5$&$26.8$&$2.1$&$27.5$&$3.1$&$24.8$&$2.3$&$33.9$&$3.3$&$27.5$&$3.5$\\

 $\text{Bactrian-X}_{\text{7B}}$
&$14.1$&$2.3$&$19.1$&$2.8$&$0.7$&$0.2$&$18.8$&$1.3$&$4.7$&$0.5$&$1.7$&$0.2$&$1.3$&$0.2$&$27.2$&$2.8$&$0.3$&$0.0$&$4.7$&$0.4$&$9.3$&$1.1$\\

 $\text{Bactrian-M}_{\text{7B}}$
&$32.9$&$6.2$&$37.2$&$6.7$&$37.6$&$5.4$&$\textbf{52.0}$&$\textbf{6.7}$&$47.7$&$6.4$&$34.2$&$3.7$&$53.4$&$4.8$&$46.3$&$6.1$&$22.1$&$1.9$&$49.0$&$5.2$&$41.2$&$5.3$\\

 $\text{X-Instruction}_{\text{7B}}$                        
&$\textbf{37.2}$&$\textbf{6.6}$&$\textbf{41.3}$&$\textbf{7.1}$&$\textbf{40.6}$&$\textbf{5.8}$&$49.7$&$6.0$&$\textbf{53.0}$&$\textbf{7.3}$&$\textbf{41.3}$&$\textbf{4.9}$&$\textbf{62.4}$&$\textbf{6.3}$&$\textbf{54.0}$&$\textbf{7.0}$&$54.7$&$5.2$&$\textbf{57.4}$&$\textbf{6.4}$&$\textbf{49.2}$&$\textbf{6.3}$\\

 $\text{X-Instruction}_{\text{7B}}^{\dagger}$  
&$35.6$&$6.2$&$37.2$&$6.9$&$35.2$&$5.0$&$47.0$&$5.9$&$48.0$&$6.4$&$35.2$&$4.2$&$52.0$&$5.3$&$41.9$&$6.0$&$\textbf{55.7}$&$\textbf{5.3}$&$51.0$&$5.6$&$43.9$&$5.7$\\

 \midrule[0.8pt]


 $\text{Bactrian-X}_{\text{13B}}$
&$18.5$&$3.3$&$21.8$&$3.5$&$3.0$&$0.4$&$25.5$&$1.9$&$6.0$&$0.7$&$1.0$&$0.2$&$1.0$&$0.3$&$29.5$&$3.6$&$1.7$&$0.2$&$5.4$&$0.5$&$11.3$&$1.5$\\
 $\text{Bactrian-M}_{\text{13B}}$        
&$37.6$&$6.6$&$39.6$&$7.5$&$41.6$&$6.0$&$49.0$&$6.8$&$50.7$&$7.3$&$38.6$&$4.4$&$58.7$&$5.5$&$50.0$&$6.7$&$28.9$&$2.6$&$51.0$&$5.2$&$44.6$&$5.9$\\

 $\text{X-Instruction}_{\text{13B}}$                        &$\textbf{47.0}$&$\textbf{7.5}$&$\textbf{50.3}$&$\textbf{8.2}$&$\textbf{53.0}$&$\textbf{7.6}$&$\textbf{53.7}$&$\textbf{7.0}$&$\textbf{57.0}$&$\textbf{7.8}$&$\textbf{68.8}$&$\textbf{7.4}$&$64.4$&$\textbf{7.1}$&$\textbf{54.4}$&$\textbf{7.1}$&$\textbf{76.8}$&$\textbf{7.6}$&$\textbf{71.1}$&$\textbf{7.5}$&$\textbf{59.7}$&$\textbf{7.5}$\\
 $\text{X-Instruction}_{\text{13B}}^{\dagger}$                       &$42.6$&$7.1$&$44.0$&$7.5$&$43.0$&$6.1$&$53.4$&$6.5$&$54.7$&$7.5$&$58.4$&$6.7$&$\textbf{65.8}$&$6.6$&$49.3$&$6.6$&$67.1$&$6.8$&$57.7$&$6.7$&$53.6$&$6.8$\\

\bottomrule[1.2pt]
\end{tabu}
\vspace{-2mm}

\caption{\label{tab:win_rates_scores_all_langs} The evaluation results of different multilingual models on 10 languages from GPT-4, where ``\textbf{W}'' represents the average win rates against ChatGPT and ``\textbf{S}'' represents the average GPT-4 score on Vicuna and WizardLM benchmarks (refer to Appendix \ref{sec:detailed_10langs} for detailed results). ${}^{\dagger}$ denotes the zero-shot evaluation results, where the X-Instruction model is prompted in the output language rather than the English prompt used in training. 
}

\vspace{-4mm}

\end{table*}

%% file: tabs/response_quality.tex
\setlength{\tabcolsep}{1.7mm}
\begin{table}[th]
	\centering
	\footnotesize
	\renewcommand\arraystretch{1.2}
	\begin{center}
		\begin{tabular}{cccc}
			\toprule[1.2pt]  
			\textbf{Model}                           & \textbf{Helpfulness}   & \textbf{Relevance}   & \textbf{Accuracy}  \\
			\midrule[0.8pt]
                $\text{Bactrian-M}_{\text{13B}}$         & $6.3_{\pm 1.2}$  & $7.2_{\pm 1.4}$  & $6.7_{\pm 1.2}$\\
                $\text{X-Instruction}_{\text{13B}}$      & $\textbf{8.1}_{\pm 0.2}$  & $\textbf{9.1}_{\pm 0.3}$  & $\textbf{8.0}_{\pm 0.2}$\\
			\bottomrule[1.2pt]
		\end{tabular}
		\caption{\label{tab:res_qe} The average evaluation results across ten languages from GPT-4 on three different views.}
	\end{center}
	\vspace{-2mm}
\end{table}

%% file: tabs/xnli_xcopa.tex
\begin{table*}[thp]

\renewcommand\arraystretch{1.1}

\centering
\scriptsize

\setlength{\tabcolsep}{1.7mm}

 \begin{tabu}{l|cccccc|cccccc|ccc|c}
 
 \toprule[1.2pt]
  \multicolumn{1}{c}{ } & \multicolumn{6}{c}{\textbf{XNLI}} & \multicolumn{6}{c}{\textbf{XCOPA}} & \multicolumn{3}{c}{\textbf{XStoryCloze}} &\\
  \cmidrule(r){2-7}  \cmidrule(r){8-13} \cmidrule(r){14-16} \noalign{\smallskip}
\multicolumn{1}{c}{\textbf{Model}}& \textbf{th}&\textbf{tr}& \textbf{vi}& \textbf{hi}&\textbf{sw}& \textbf{ur}& \textbf{id}&\textbf{th}& \textbf{tr}&\textbf{vi} & \textbf{sw}&\textbf{ta} &\textbf{id} & \textbf{hi}&\textbf{sw}  &\multicolumn{1}{c}{\textbf{Avg}}\\

   \midrule[0.8pt]

  $\text{LLaMA 2}_{\text{7B}}$            
  &$42.3$   & $41.6$   & $44.4$   & $44.2$   & $35.4$   & $40.1$   & $58.2$   & $57.2$   & $53.0$   & $57.0$   & $51.8$   & $54.2$   & $59.6$   & $53.6$   & $50.4$   & $49.5$\\

  $\text{Bactrian-M}_{\text{7B}}$       
  &$\textbf{42.8}$   & $42.1$   & $44.5$   & $43.9$   & $40.6$   & $41.5$   & $60.2$   & $\textbf{57.4}$   & $53.4$   & $59.4$   & $53.2$   & $55.0$   & $62.9$   & $55.1$   & $54.2$   & $51.1$\\

  $\text{X-Instruction}_{\text{7B}}$       
  & $42.4$   & $\textbf{42.7}$   & $\textbf{44.8}$   & $\textbf{45.6}$   & $\textbf{41.5}$   & $\textbf{42.0}$   & $\textbf{61.4}$   & $\textbf{57.4}$   & $\textbf{55.6}$   & $\textbf{60.4}$   & $\textbf{56.0}$   & $\textbf{57.4}$   & $\textbf{64.2}$   & $\textbf{59.8}$   & $\textbf{56.6}$   & $\textbf{52.5}$\\

  \midrule[0.8pt]
  

$\text{LLaMA 2}_{\text{13B}}$   
& $41.2$   & $44.0$   & $45.9$   & $44.8$   & $35.5$   & $41.7$   & $61.6$   & $56.2$   & $55.0$   & $60.6$   & $52.6$   & $52.6$   & $64.1$   & $55.4$   & $51.5$   & $50.8$\\

 $\text{Bactrian-M}_{\text{13B}}$       
 & $43.5$   & $44.3$   & $\textbf{46.6}$   & $45.4$   & $41.2$   & $43.4$   & $\textbf{62.8}$   & $56.8$   & $55.8$   & $62.2$   & $53.8$   & $53.8$   & $66.0$   & $59.2$   & $57.4$   & $52.8$ \\

  $\text{X-Instruction}_{\text{13B}}$       
  & $\textbf{44.5}$   & $\textbf{44.9}$   & $\textbf{46.6}$   & $\textbf{46.4}$   & $\textbf{41.9}$   & $\textbf{45.2}$   & $62.0$   & $\textbf{57.0}$   & $\textbf{56.6}$   & $\textbf{62.8}$   & $\textbf{58.8}$   & $\textbf{56.6}$   & $\textbf{67.0}$   & $\textbf{61.4}$   & $\textbf{58.2}$   & $\textbf{54.0}$ \\

 \specialrule{0em}{0pt}{0pt}

\bottomrule[1.2pt]
\end{tabu}
\vspace{-2mm}

\caption{\label{tab:more_benchmark} Zero-shot in-context learning performance on NLI and Reasoning datasets across languages. Following \citet{lin-etal-2022-shot}, the prompt template is written in English for all languages evaluated. 
}


\end{table*}

%% file: tabs/seed_statistic.tex
\setlength{\tabcolsep}{0.8mm}
\begin{table}[ht]
	\centering
	\footnotesize
	\renewcommand\arraystretch{1}
	\begin{center}
		\begin{tabular}{ccccc}
			\toprule[1.2pt]  
			\textbf{Lang} & \textbf{\#Sample}   & \textbf{Instruction Length(en)}   & \textbf{Output Length}  \\
			\midrule[0.8pt]
                fi        & 3,009          & $143.0_{\pm 211.3}$  & $1073.7_{\pm 792.6}$  \\
                id        & 3,003          & $141.5_{\pm 226.2}$  & $1145.1_{\pm 829.6}$  \\
                th        & 3,166          & $138.1_{\pm 219.7}$  & $902.7_{\pm 691.8}$  \\
                tr        & 3,009          & $144.0_{\pm 230.7}$  & $1078.8_{\pm 793.4}$  \\
                vi        & 3,040          & $142.7_{\pm 227.4}$  & $1077.6_{\pm 798.8}$  \\
                bn${}^{\star}$        & 3,000          & $142.9_{\pm 225.3}$  & $1064.6_{\pm 771.9}$  \\
                hi${}^{\star}$        & 3,000          & $141.4_{\pm 212.2}$  & $1070.2_{\pm 780.2}$  \\
                sw${}^{\star}$        & 3,000          & $142.1_{\pm 225.3}$  & $1085.2_{\pm 786.9}$  \\
                ta${}^{\star}$        & 3,000          & $143.7_{\pm 227.6}$  & $1184.5_{\pm 869.1}$  \\
                ur${}^{\star}$        & 3,000          & $143.4_{\pm 230.1}$  & $1027.6_{\pm 755.4}$  \\
			\bottomrule[1.2pt]
		\end{tabular}
		\caption{\label{tab:seed_statis} Statistic of cross-lingual instruction tuning seed data with English instruction. ${}^{\star}$ indicates the low-resource language.}
	\end{center}
	\vspace{-2mm}
\end{table}

%% file: tabs/datasets_details.tex
\setlength{\tabcolsep}{2mm}
\begin{table}[h]
	\centering
	\footnotesize
	\vspace{-0.2cm}
	\renewcommand\arraystretch{1.2}
	\begin{center}
		\begin{tabular}{lcccc}
			\toprule[1.2pt]  
			\textbf{Datasets}   & \textbf{\#Samples} & \textbf{Category}  \\
			\midrule[0.8pt]
                Vicuna &  80 & \checkmark  \\
                Koala &  180 & -  \\
                WizardLLM &  218 & \checkmark  \\
                LIMA &  300 & -  \\
			\bottomrule[1.2pt]
		\end{tabular}
		\caption{\label{tab:datasets_details} The details of four benchmarks.}
	\end{center}
	\vspace{-2mm}
\end{table}

%% file: tabs/sft_statistic.tex
\setlength{\tabcolsep}{0.8mm}
\begin{table}[ht]
	\centering
	\footnotesize
	\renewcommand\arraystretch{1}
	\begin{center}
		\begin{tabular}{ccccc}
			\toprule[1.2pt]  
			\textbf{Dataset} & \textbf{Instruction Len.}   & \textbf{Input Len.}   & \textbf{Output Len.}  \\
			\midrule[0.8pt]
                Alpaca-MT     & $61.7_{\pm 29.9}$          & $62.4_{\pm 89.4}$  & $375.8_{\pm 413.1}$  \\
                Bactrian-X    & $62.6_{\pm 64.6}$          & $269.4_{\pm 772.6}$  & $514.3_{\pm 510.6}$  \\
			\bottomrule[1.2pt]
		\end{tabular}
		\caption{\label{tab:sft_statis} The average length of baseline multilingual instruction tuning datasets across ten languages.}
	\end{center}
	\vspace{-2mm}
\end{table}

%% file: tabs/detailed_results_for_two_benchmarks.tex
\begin{table*}[thp]

\renewcommand\arraystretch{1.1}

\centering
\scriptsize

\setlength{\tabcolsep}{0.6mm}

 \begin{tabu}{l|cc|cc|cc|cc|cc|cc|cc|cc|cc|cc|cc}
 
 \toprule[1.2pt]
  \multicolumn{1}{c}{ } & \multicolumn{10}{c}{\textbf{Medium}} & \multicolumn{10}{c}{\textbf{Low}} &\\
  
  \cmidrule(r){2-11}  \cmidrule(r){12-21} \noalign{\smallskip}
  
  \multicolumn{1}{c}{ } & \multicolumn{2}{c}{\textbf{fi}} & \multicolumn{2}{c}{\textbf{id}} & \multicolumn{2}{c}{\textbf{th}} & \multicolumn{2}{c}{\textbf{tr}} & \multicolumn{2}{c}{\textbf{vi}} & \multicolumn{2}{c}{\textbf{bn}} & \multicolumn{2}{c}{\textbf{hi}} & \multicolumn{2}{c}{\textbf{sw}} & \multicolumn{2}{c}{\textbf{ta}} & \multicolumn{2}{c}{\textbf{ur}} & \multicolumn{2}{c}{\textbf{Avg}} \\
  \cmidrule(r){2-3} \cmidrule(r){4-5} \cmidrule(r){6-7} \cmidrule(r){8-9} \cmidrule(r){10-11} \cmidrule(r){12-13} \cmidrule(r){14-15} \cmidrule(r){16-17} \cmidrule(r){18-19} \cmidrule(r){20-21} \cmidrule(r){22-23} \noalign{\smallskip}
  
\multicolumn{1}{c}{\textbf{Model}}& \textbf{W}&\textbf{S}& \textbf{W}&\textbf{S}& \textbf{W}&\textbf{S}& \textbf{W}&\textbf{S}& \textbf{W}&\textbf{S}& \textbf{W}&\textbf{S} & \textbf{W}&\textbf{S} & \textbf{W}&\textbf{S} & \textbf{W}&\textbf{S} & \textbf{W}&\textbf{S} & \textbf{W}&\textbf{S}\\

\midrule[0.8pt]

\multicolumn{1}{c}{$\sim$\textsc{7B Model}} &\multicolumn{20}{c}{\textsc{Vicuna dataset}}\\
\midrule[0.8pt]

$\text{Alpaca-MT}_{\text{7B}}$
&$31.3$&$6.2$&$33.8$&$6.9$&$21.3$&$3.1$&$50.0$&$4.5$&$43.8$&$6.7$&$20.0$&$2.0$&$30.0$&$1.9$&$41.3$&$4.7$&$25.0$&$1.9$&$37.5$&$3.6$&$33.4$&$4.15$\\

 $\text{Bactrian-X}_{\text{7B}}$
&$22.5$&$3.3$&$21.3$&$3.3$&$0.0$&$0.1$&$32.5$&$1.8$&$7.5$&$0.3$&$1.3$&$0.2$&$1.3$&$0.1$&$42.5$&$4.3$&$0.0$&$0.0$&$6.3$&$0.5$&$13.52$&$1.39$\\

 $\text{Bactrian-M}_{\text{7B}}$
&$41.3$&$7.4$&$46.3$&$7.7$&$45.0$&$6.9$&$58.8$&$6.5$&$55.0$&$7.7$&$45.0$&$4.8$&$72.5$&$6.6$&$56.3$&$8.1$&$42.5$&$2.7$&$57.5$&$6.9$&$52.02$&$6.53$\\

 $\text{X-Instruction}_{\text{7B}}$                        
&$42.5$&$7.9$&$50.0$&$8.3$&$51.3$&$7.5$&$67.5$&$7.4$&$65.0$&$8.5$&$61.3$&$7.3$&$77.5$&$8.0$&$60.0$&$8.0$&$80.0$&$7.1$&$60.0$&$7.0$&$61.51$&$7.70$\\

 $\text{X-Instruction}_{\text{7B}}^{\dagger}$  
&$42.5$&$7.6$&$46.3$&$7.9$&$43.8$&$6.4$&$57.5$&$6.7$&$65.0$&$8.1$&$42.5$&$5.2$&$70.0$&$6.7$&$42.5$&$6.4$&$67.5$&$5.9$&$58.8$&$6.6$&$53.64$&$6.75$\\

\midrule[0.8pt]
\multicolumn{1}{c}{ } &\multicolumn{20}{c}{\textsc{WizardLM dataset}}\\
\midrule[0.8pt]

$\text{Alpaca-MT}_{\text{7B}}$
&$19.7$&$4.2$&$21.1$&$4.8$&$24.3$&$3.0$&$28.9$&$2.9$&$32.6$&$4.5$&$20.6$&$2.7$&$25.7$&$2.2$&$22.5$&$2.5$&$24.8$&$2.4$&$32.6$&$3.2$&$25.28$&$3.24$\\

 $\text{Bactrian-X}_{\text{7B}}$
&$11.0$&$1.9$&$18.3$&$2.6$&$0.9$&$0.3$&$13.8$&$1.2$&$3.7$&$0.5$&$1.8$&$0.2$&$1.4$&$0.2$&$21.6$&$2.2$&$0.5$&$0.1$&$4.1$&$0.3$&$7.71$&$0.95$\\

 $\text{Bactrian-M}_{\text{7B}}$
&$29.8$&$5.7$&$33.9$&$6.3$&$34.9$&$4.8$&$49.5$&$6.7$&$45.0$&$6.0$&$30.3$&$3.2$&$46.3$&$4.1$&$42.7$&$5.4$&$14.7$&$1.5$&$45.9$&$4.6$&$37.30$&$4.83$\\

 $\text{X-Instruction}_{\text{7B}}$                        
&$35.3$&$6.2$&$38.1$&$6.6$&$36.7$&$5.1$&$43.1$&$5.5$&$48.6$&$6.8$&$33.9$&$4.0$&$56.9$&$5.7$&$51.8$&$6.6$&$45.4$&$4.5$&$56.4$&$6.2$&$44.62$&$5.72$\\

 $\text{X-Instruction}_{\text{7B}}^{\dagger}$  
&$33.0$&$5.7$&$33.9$&$6.5$&$32.1$&$4.5$&$43.1$&$5.6$&$41.7$&$5.9$&$32.6$&$3.8$&$45.4$&$4.8$&$41.7$&$5.8$&$51.4$&$5.0$&$48.2$&$5.3$&$40.31$&$5.29$\\

 \midrule[0.8pt]
\multicolumn{1}{c}{$\sim$\textsc{13B Model}} &\multicolumn{20}{c}{\textsc{Vicuna dataset}} & \multicolumn{1}{c}{ }\\
\midrule[0.8pt]

 $\text{Bactrian-X}_{\text{13B}}$
&$22.5$&$4.5$&$27.5$&$4.0$&$1.3$&$0.2$&$46.3$&$2.3$&$3.8$&$0.4$&$0.0$&$0.0$&$1.3$&$0.1$&$41.3$&$5.0$&$0.0$&$0.1$&$5.0$&$0.4$&$14.90$&$1.7$\\

 $\text{Bactrian-M}_{\text{13B}}$        &$48.8$&$7.7$&$47.5$&$8.5$&$46.3$&$7.4$&$66.3$&$8.1$&$61.3$&$8.2$&$48.8$&$5.6$&$75.0$&$7.0$&$58.8$&$7.7$&$42.5$&$3.7$&$56.3$&$6.6$&$55.16$&$7.05$\\

 $\text{X-Instruction}_{\text{13B}}$                        &$52.5$&$8.4$&$58.8$&$9.0$&$61.3$&$8.7$&$67.5$&$8.2$&$65.0$&$8.7$&$85.0$&$8.8$&$86.3$&$9.0$&$65.0$&$8.5$&$95.0$&$9.1$&$78.8$&$8.7$&$71.52$&$8.71$\\

 $\text{X-Instruction}_{\text{13B}}^{\dagger}$                       &$50.0$&$8.3$&$52.5$&$8.6$&$42.5$&$6.9$&$61.3$&$7.0$&$65.0$&$8.4$&$67.5$&$7.7$&$85.0$&$8.2$&$52.5$&$7.2$&$82.5$&$7.8$&$58.8$&$7.7$&$61.76$&$7.78$\\

 \midrule[0.8pt]
\multicolumn{1}{c}{ } &\multicolumn{20}{c}{\textsc{WizardLM dataset}} & \multicolumn{1}{c}{ }\\
\midrule[0.8pt]

 $\text{Bactrian-X}_{\text{13B}}$
&$17.0$&$2.9$&$19.7$&$3.3$&$3.7$&$0.5$&$17.9$&$1.8$&$6.9$&$0.8$&$1.4$&$0.2$&$0.9$&$0.3$&$25.2$&$3.1$&$2.3$&$0.3$&$5.5$&$0.6$&$10.05$&$1.38$\\

 $\text{Bactrian-M}_{\text{13B}}$        &$33.5$&$6.2$&$36.7$&$7.1$&$39.9$&$5.5$&$42.7$&$6.4$&$46.8$&$7.0$&$34.9$&$3.9$&$52.8$&$4.9$&$46.8$&$6.4$&$23.9$&$2.2$&$49.1$&$4.7$&$40.71$&$5.43$\\

 $\text{X-Instruction}_{\text{13B}}$                        &$45.0$&$7.1$&$47.2$&$8.0$&$50.0$&$7.2$&$48.6$&$6.6$&$54.1$&$7.5$&$62.8$&$6.9$&$56.4$&$6.4$&$50.5$&$6.6$&$70.2$&$7.0$&$68.3$&$7.1$&$55.31$&$7.04$\\

 $\text{X-Instruction}_{\text{13B}}^{\dagger}$                       &$39.9$&$6.7$&$40.8$&$7.1$&$43.1$&$5.8$&$50.5$&$6.3$&$50.9$&$7.2$&$55.0$&$6.3$&$58.7$&$6.0$&$48.2$&$6.3$&$61.5$&$6.5$&$57.3$&$6.3$&$50.59$&$6.45$\\

\bottomrule[1.2pt]
\end{tabu}
\vspace{-2mm}

\caption{\label{tab:all_langs_detail} The detailed results of different multilingual models on 10 languages from GPT-4, where ``\textbf{W}'' represents the win rates to ChatGPT and ``\textbf{S}'' represents the GPT-4 score. ${}^{\dagger}$ denotes the zero-shot evaluation results.
}

\vspace{-2mm}

\end{table*}

%% file: tabs/seed_improve.tex
\setlength{\tabcolsep}{2mm}
\begin{table}[ht]
	\centering
	\footnotesize
	\renewcommand\arraystretch{1.2}
	\begin{center}
		\begin{tabular}{lcccc}
			\toprule[1.2pt]  
			\textbf{Category}   & $\text{Seed Model}_{\text{7B}}$ & $\text{X-Instruction}_{\text{7B}}$  \\
			\midrule[0.8pt]
                Counterfactual  & 4.7  &9.3 (+4.7) \\
                Common-sense    & 5.0  &8.3 (+3.3) \\
                Roleplay        & 1.7  &5.0 (+3.3) \\
                Coding          & 2.0  &5.0 (+3.0) \\
                Knowledge       & 4.0  &6.0 (+2.0) \\
                Writing         & 3.3  &5.3 (+2.0) \\
                Generic         & 7.0  &7.7 (+0.7) \\
                Math            & 0.7  &1.3 (+0.6) \\
                Fermi           & 1.7  &2.0 (+0.3) \\
			\midrule[0.8pt]
                All             & 30.0 &\textbf{50.0} (+20.0) \\
			\bottomrule[1.2pt]
		\end{tabular}
		\caption{\label{tab:seed_improve} The average number of questions won to ChatGPT on the Vicuna dataset in 3 languages.}
	\end{center}
	\vspace{-4mm}
\end{table}

%% file: tabs/language_codes.tex
\begin{table}[htp]


    \setlength{\tabcolsep}{2mm}
	\centering
	\renewcommand\arraystretch{1.25}
	\begin{center}
		\begin{tabular}{ccc}
			\toprule[1.2pt]  
               \multicolumn{1}{c}{\textbf{ISO 639-1}} & \multicolumn{1}{c}{\textbf{Language}}    &\multicolumn{1}{c}{\textbf{Family}} \\
                \midrule[0.8pt]
                bn      &Bengali      &Indo-European\\
                en      &English      &Indo-European\\
                fi      &Finnish      &Uralic\\
                hi      &Hindi      &Indo-European\\
                id      &Indonesian      &Austronesian\\
                ta      &Tamil      &Dravidian\\
                th      &Thai      &Kra-Dai\\
                tr      &Turkish      &Turkic\\
                vi      &Vietnamese      &Austroasiatic\\
                sw      &Swahili      &Niger-Congo\\
                ur      &Urdu      &Indo-European\\
			\bottomrule[1.2pt]
		\end{tabular}
	\end{center}

\caption{\label{tab:lang_codes} Details of Language codes in this work. 
}

\end{table}

%% file: acl_latex.bbl
\begin{thebibliography}{44}
\expandafter\ifx\csname natexlab\endcsname\relax\def\natexlab#1{#1}\fi

\bibitem[{Asai et~al.(2023)Asai, Kudugunta, Yu, Blevins, Gonen, Reid, Tsvetkov, Ruder, and Hajishirzi}]{asai2023buffet}
Akari Asai, Sneha Kudugunta, Xinyan~Velocity Yu, Terra Blevins, Hila Gonen, Machel Reid, Yulia Tsvetkov, Sebastian Ruder, and Hannaneh Hajishirzi. 2023.
\newblock Buffet: Benchmarking large language models for few-shot cross-lingual transfer.
\newblock \emph{arXiv preprint arXiv:2305.14857}.

\bibitem[{Bai et~al.(2022)Bai, Kadavath, Kundu, Askell, Kernion, Jones, Chen, Goldie, Mirhoseini, McKinnon et~al.}]{bai2022constitutional}
Yuntao Bai, Saurav Kadavath, Sandipan Kundu, Amanda Askell, Jackson Kernion, Andy Jones, Anna Chen, Anna Goldie, Azalia Mirhoseini, Cameron McKinnon, et~al. 2022.
\newblock Constitutional ai: Harmlessness from ai feedback.
\newblock \emph{arXiv preprint arXiv:2212.08073}.

\bibitem[{Chen et~al.(2023)Chen, Li, Yan, Wang, Gunaratna, Yadav, Tang, Srinivasan, Zhou, Huang et~al.}]{chen2023alpagasus}
Lichang Chen, Shiyang Li, Jun Yan, Hai Wang, Kalpa Gunaratna, Vikas Yadav, Zheng Tang, Vijay Srinivasan, Tianyi Zhou, Heng Huang, et~al. 2023.
\newblock Alpagasus: Training a better alpaca with fewer data.
\newblock \emph{arXiv preprint arXiv:2307.08701}.

\bibitem[{Chung et~al.(2022)Chung, Hou, Longpre, Zoph, Tay, Fedus, Li, Wang, Dehghani, Brahma et~al.}]{chung2022scaling}
Hyung~Won Chung, Le~Hou, Shayne Longpre, Barret Zoph, Yi~Tay, William Fedus, Yunxuan Li, Xuezhi Wang, Mostafa Dehghani, Siddhartha Brahma, et~al. 2022.
\newblock Scaling instruction-finetuned language models.
\newblock \emph{arXiv preprint arXiv:2210.11416}.

\bibitem[{Conneau et~al.(2020)Conneau, Khandelwal, Goyal, Chaudhary, Wenzek, Guzm{\'a}n, Grave, Ott, Zettlemoyer, and Stoyanov}]{conneau-etal-2020-unsupervised}
Alexis Conneau, Kartikay Khandelwal, Naman Goyal, Vishrav Chaudhary, Guillaume Wenzek, Francisco Guzm{\'a}n, Edouard Grave, Myle Ott, Luke Zettlemoyer, and Veselin Stoyanov. 2020.
\newblock \href {https://doi.org/10.18653/v1/2020.acl-main.747} {Unsupervised cross-lingual representation learning at scale}.
\newblock In \emph{Proceedings of the 58th Annual Meeting of the Association for Computational Linguistics}, pages 8440--8451, Online. Association for Computational Linguistics.

\bibitem[{Conneau et~al.(2018)Conneau, Rinott, Lample, Williams, Bowman, Schwenk, and Stoyanov}]{conneau-etal-2018-xnli}
Alexis Conneau, Ruty Rinott, Guillaume Lample, Adina Williams, Samuel Bowman, Holger Schwenk, and Veselin Stoyanov. 2018.
\newblock \href {https://doi.org/10.18653/v1/D18-1269} {{XNLI}: Evaluating cross-lingual sentence representations}.
\newblock In \emph{Proceedings of the 2018 Conference on Empirical Methods in Natural Language Processing}, pages 2475--2485, Brussels, Belgium. Association for Computational Linguistics.

\bibitem[{Geng et~al.(2023)Geng, Gudibande, Hao, Eric, Pieter, Sergey, and Dawn}]{Geng2023koala}
Xinyang Geng, Arnav Gudibande, Liu Hao, Wallace Eric, Abbeel Pieter, Levine Sergey, and Song Dawn. 2023.
\newblock \href {https://bair.berkeley.edu/blog/2023/04/03/koala/} {Koala: A dialogue model for academic research}.
\newblock \emph{Blog post}.

\bibitem[{Holtzman et~al.(2020)Holtzman, Buys, Du, Forbes, and Choi}]{Holtzman2020topp}
Ari Holtzman, Jan Buys, Li~Du, Maxwell Forbes, and Yejin Choi. 2020.
\newblock \href {https://openreview.net/forum?id=rygGQyrFvH} {The curious case of neural text degeneration}.
\newblock In \emph{International Conference on Learning Representations}.

\bibitem[{Kitaev et~al.(2019)Kitaev, Cao, and Klein}]{kitaev-etal-2019-multilingual}
Nikita Kitaev, Steven Cao, and Dan Klein. 2019.
\newblock \href {https://doi.org/10.18653/v1/P19-1340} {Multilingual constituency parsing with self-attention and pre-training}.
\newblock In \emph{Proceedings of the 57th Annual Meeting of the Association for Computational Linguistics}, pages 3499--3505, Florence, Italy. Association for Computational Linguistics.

\bibitem[{Kitaev and Klein(2018)}]{kitaev-klein-2018-constituency}
Nikita Kitaev and Dan Klein. 2018.
\newblock \href {https://doi.org/10.18653/v1/P18-1249} {Constituency parsing with a self-attentive encoder}.
\newblock In \emph{Proceedings of the 56th Annual Meeting of the Association for Computational Linguistics (Volume 1: Long Papers)}, pages 2676--2686, Melbourne, Australia. Association for Computational Linguistics.

\bibitem[{K{\"o}pf et~al.(2023)K{\"o}pf, Kilcher, von R{\"u}tte, Anagnostidis, Tam, Stevens, Barhoum, Duc, Stanley, Nagyfi et~al.}]{kopf2023openassistant}
Andreas K{\"o}pf, Yannic Kilcher, Dimitri von R{\"u}tte, Sotiris Anagnostidis, Zhi-Rui Tam, Keith Stevens, Abdullah Barhoum, Nguyen~Minh Duc, Oliver Stanley, Rich{\'a}rd Nagyfi, et~al. 2023.
\newblock Openassistant conversations--democratizing large language model alignment.
\newblock \emph{arXiv preprint arXiv:2304.07327}.

\bibitem[{Li et~al.(2023{\natexlab{a}})Li, Wang, Zhang, and Zong}]{li2023align}
Chong Li, Shaonan Wang, Jiajun Zhang, and Chengqing Zong. 2023{\natexlab{a}}.
\newblock Align after pre-train: Improving multilingual generative models with cross-lingual alignment.
\newblock \emph{arXiv preprint arXiv:2311.08089}.

\bibitem[{Li et~al.(2023{\natexlab{b}})Li, Koto, Wu, Aji, and Baldwin}]{li2023bactrian}
Haonan Li, Fajri Koto, Minghao Wu, Alham~Fikri Aji, and Timothy Baldwin. 2023{\natexlab{b}}.
\newblock Bactrian-x: A multilingual replicable instruction-following model with low-rank adaptation.
\newblock \emph{arXiv preprint arXiv:2305.15011}.

\bibitem[{Li et~al.(2023{\natexlab{c}})Li, Yu, Zhou, Schick, Zettlemoyer, Levy, Weston, and Lewis}]{li2023self}
Xian Li, Ping Yu, Chunting Zhou, Timo Schick, Luke Zettlemoyer, Omer Levy, Jason Weston, and Mike Lewis. 2023{\natexlab{c}}.
\newblock Self-alignment with instruction backtranslation.
\newblock \emph{arXiv preprint arXiv:2308.06259}.

\bibitem[{Li et~al.(2023{\natexlab{d}})Li, Zhang, Dubois, Taori, Gulrajani, Guestrin, Liang, and Hashimoto}]{alpaca_eval}
Xuechen Li, Tianyi Zhang, Yann Dubois, Rohan Taori, Ishaan Gulrajani, Carlos Guestrin, Percy Liang, and Tatsunori~B. Hashimoto. 2023{\natexlab{d}}.
\newblock Alpacaeval: An automatic evaluator of instruction-following models.
\newblock \url{https://github.com/tatsu-lab/alpaca_eval}.

\bibitem[{Lin et~al.(2022)Lin, Mihaylov, Artetxe, Wang, Chen, Simig, Ott, Goyal, Bhosale, Du, Pasunuru, Shleifer, Koura, Chaudhary, O{'}Horo, Wang, Zettlemoyer, Kozareva, Diab, Stoyanov, and Li}]{lin-etal-2022-shot}
Xi~Victoria Lin, Todor Mihaylov, Mikel Artetxe, Tianlu Wang, Shuohui Chen, Daniel Simig, Myle Ott, Naman Goyal, Shruti Bhosale, Jingfei Du, Ramakanth Pasunuru, Sam Shleifer, Punit~Singh Koura, Vishrav Chaudhary, Brian O{'}Horo, Jeff Wang, Luke Zettlemoyer, Zornitsa Kozareva, Mona Diab, Veselin Stoyanov, and Xian Li. 2022.
\newblock \href {https://doi.org/10.18653/v1/2022.emnlp-main.616} {Few-shot learning with multilingual generative language models}.
\newblock In \emph{Proceedings of the 2022 Conference on Empirical Methods in Natural Language Processing}, pages 9019--9052, Abu Dhabi, United Arab Emirates. Association for Computational Linguistics.

\bibitem[{Liu et~al.(2023{\natexlab{a}})Liu, Koto, Baldwin, and Gurevych}]{liu2023multilingual}
Chen~Cecilia Liu, Fajri Koto, Timothy Baldwin, and Iryna Gurevych. 2023{\natexlab{a}}.
\newblock Are multilingual llms culturally-diverse reasoners? an investigation into multicultural proverbs and sayings.
\newblock \emph{arXiv preprint arXiv:2309.08591}.

\bibitem[{Liu et~al.(2023{\natexlab{b}})Liu, Iter, Xu, Wang, Xu, and Zhu}]{liu-etal-2023-g}
Yang Liu, Dan Iter, Yichong Xu, Shuohang Wang, Ruochen Xu, and Chenguang Zhu. 2023{\natexlab{b}}.
\newblock \href {https://doi.org/10.18653/v1/2023.emnlp-main.153} {{G}-eval: {NLG} evaluation using gpt-4 with better human alignment}.
\newblock In \emph{Proceedings of the 2023 Conference on Empirical Methods in Natural Language Processing}, pages 2511--2522, Singapore. Association for Computational Linguistics.

\bibitem[{Loshchilov and Hutter(2019)}]{loshchilov2019adamw}
Ilya Loshchilov and Frank Hutter. 2019.
\newblock \href {https://openreview.net/forum?id=Bkg6RiCqY7} {Decoupled weight decay regularization}.
\newblock In \emph{International Conference on Learning Representations}.

\bibitem[{Micikevicius et~al.(2018)Micikevicius, Narang, Alben, Diamos, Elsen, Garcia, Ginsburg, Houston, Kuchaiev, Venkatesh et~al.}]{micikevicius2018mixed}
Paulius Micikevicius, Sharan Narang, Jonah Alben, Gregory Diamos, Erich Elsen, David Garcia, Boris Ginsburg, Michael Houston, Oleksii Kuchaiev, Ganesh Venkatesh, et~al. 2018.
\newblock Mixed precision training.
\newblock In \emph{International Conference on Learning Representations}.

\bibitem[{Mishra et~al.(2022)Mishra, Khashabi, Baral, and Hajishirzi}]{mishra-etal-2022-cross}
Swaroop Mishra, Daniel Khashabi, Chitta Baral, and Hannaneh Hajishirzi. 2022.
\newblock \href {https://doi.org/10.18653/v1/2022.acl-long.244} {Cross-task generalization via natural language crowdsourcing instructions}.
\newblock In \emph{Proceedings of the 60th Annual Meeting of the Association for Computational Linguistics (Volume 1: Long Papers)}, pages 3470--3487, Dublin, Ireland. Association for Computational Linguistics.

\bibitem[{Muennighoff et~al.(2023)Muennighoff, Wang, Sutawika, Roberts, Biderman, Le~Scao, Bari, Shen, Yong, Schoelkopf, Tang, Radev, Aji, Almubarak, Albanie, Alyafeai, Webson, Raff, and Raffel}]{muennighoff-etal-2023-crosslingual}
Niklas Muennighoff, Thomas Wang, Lintang Sutawika, Adam Roberts, Stella Biderman, Teven Le~Scao, M~Saiful Bari, Sheng Shen, Zheng~Xin Yong, Hailey Schoelkopf, Xiangru Tang, Dragomir Radev, Alham~Fikri Aji, Khalid Almubarak, Samuel Albanie, Zaid Alyafeai, Albert Webson, Edward Raff, and Colin Raffel. 2023.
\newblock \href {https://doi.org/10.18653/v1/2023.acl-long.891} {Crosslingual generalization through multitask finetuning}.
\newblock In \emph{Proceedings of the 61st Annual Meeting of the Association for Computational Linguistics (Volume 1: Long Papers)}, pages 15991--16111, Toronto, Canada. Association for Computational Linguistics.

\bibitem[{Nguyen et~al.(2023)Nguyen, Van~Nguyen, Lai, Man, Ngo, Dernoncourt, Rossi, and Nguyen}]{nguyen2023culturax}
Thuat Nguyen, Chien Van~Nguyen, Viet~Dac Lai, Hieu Man, Nghia~Trung Ngo, Franck Dernoncourt, Ryan~A Rossi, and Thien~Huu Nguyen. 2023.
\newblock Culturax: A cleaned, enormous, and multilingual dataset for large language models in 167 languages.
\newblock \emph{arXiv preprint arXiv:2309.09400}.

\bibitem[{OpenAI(2022)}]{openai2022chatgpt}
OpenAI. 2022.
\newblock \href {https://openai.com/blog/chatgpt} {Introducing chatgpt}.
\newblock \emph{OpenAI blog}.

\bibitem[{OpenAI(2023)}]{openai2023gpt4}
OpenAI. 2023.
\newblock \href {https://arxiv.org/abs/2303.08774} {Gpt-4 technical report}.
\newblock \emph{arXiv preprint arXiv:2303.08774}.

\bibitem[{Ouyang et~al.(2022{\natexlab{a}})Ouyang, Wu, Jiang, Almeida, Wainwright, Mishkin, Zhang, Agarwal, Slama, Ray, Schulman, Hilton, Kelton, Miller, Simens, Askell, Welinder, Christiano, Leike, and Lowe}]{long2022instructgpt}
Long Ouyang, Jeffrey Wu, Xu~Jiang, Diogo Almeida, Carroll Wainwright, Pamela Mishkin, Chong Zhang, Sandhini Agarwal, Katarina Slama, Alex Ray, John Schulman, Jacob Hilton, Fraser Kelton, Luke Miller, Maddie Simens, Amanda Askell, Peter Welinder, Paul~F Christiano, Jan Leike, and Ryan Lowe. 2022{\natexlab{a}}.
\newblock \href {https://proceedings.neurips.cc/paper_files/paper/2022/file/b1efde53be364a73914f58805a001731-Paper-Conference.pdf} {Training language models to follow instructions with human feedback}.
\newblock In \emph{Advances in Neural Information Processing Systems}, volume~35, pages 27730--27744. Curran Associates, Inc.

\bibitem[{Ouyang et~al.(2022{\natexlab{b}})Ouyang, Wu, Jiang, Almeida, Wainwright, Mishkin, Zhang, Agarwal, Slama, Ray et~al.}]{ouyang2022training}
Long Ouyang, Jeffrey Wu, Xu~Jiang, Diogo Almeida, Carroll Wainwright, Pamela Mishkin, Chong Zhang, Sandhini Agarwal, Katarina Slama, Alex Ray, et~al. 2022{\natexlab{b}}.
\newblock Training language models to follow instructions with human feedback.
\newblock \emph{Advances in Neural Information Processing Systems}, 35:27730--27744.

\bibitem[{Ponti et~al.(2020)Ponti, Glava{\v{s}}, Majewska, Liu, Vuli{\'c}, and Korhonen}]{ponti-etal-2020-xcopa}
Edoardo~Maria Ponti, Goran Glava{\v{s}}, Olga Majewska, Qianchu Liu, Ivan Vuli{\'c}, and Anna Korhonen. 2020.
\newblock \href {https://doi.org/10.18653/v1/2020.emnlp-main.185} {{XCOPA}: A multilingual dataset for causal commonsense reasoning}.
\newblock In \emph{Proceedings of the 2020 Conference on Empirical Methods in Natural Language Processing (EMNLP)}, pages 2362--2376, Online. Association for Computational Linguistics.

\bibitem[{Rasley et~al.(2020)Rasley, Rajbhandari, Ruwase, and He}]{rasley2020deepspeed}
Jeff Rasley, Samyam Rajbhandari, Olatunji Ruwase, and Yuxiong He. 2020.
\newblock \href {https://doi.org/10.1145/3394486.3406703} {Deepspeed: System optimizations enable training deep learning models with over 100 billion parameters}.
\newblock In \emph{Proceedings of the 26th ACM SIGKDD International Conference on Knowledge Discovery \& Data Mining}, KDD '20, page 3505–3506, New York, NY, USA. Association for Computing Machinery.

\bibitem[{Song et~al.(2020)Song, Tan, Qin, Lu, and Liu}]{song2020mpnet}
Kaitao Song, Xu~Tan, Tao Qin, Jianfeng Lu, and Tie-Yan Liu. 2020.
\newblock Mpnet: Masked and permuted pre-training for language understanding.
\newblock \emph{Advances in Neural Information Processing Systems}, 33:16857--16867.

\bibitem[{Su{\'a}rez et~al.(2019)Su{\'a}rez, Sagot, and Romary}]{suarez2019asynchronous}
Pedro Javier~Ortiz Su{\'a}rez, Beno{\^\i}t Sagot, and Laurent Romary. 2019.
\newblock Asynchronous pipeline for processing huge corpora on medium to low resource infrastructures.
\newblock In \emph{7th Workshop on the Challenges in the Management of Large Corpora (CMLC-7)}. Leibniz-Institut f{\"u}r Deutsche Sprache.

\bibitem[{Sun et~al.(2023)Sun, Shen, Zhou, Zhang, Chen, Cox, Yang, and Gan}]{sun2023principle}
Zhiqing Sun, Yikang Shen, Qinhong Zhou, Hongxin Zhang, Zhenfang Chen, David Cox, Yiming Yang, and Chuang Gan. 2023.
\newblock Principle-driven self-alignment of language models from scratch with minimal human supervision.
\newblock \emph{arXiv preprint arXiv:2305.03047}.

\bibitem[{Taori et~al.(2023)Taori, Gulrajani, Zhang, Dubois, Li, Guestrin, Liang, and Hashimoto}]{taori2023alpaca}
Rohan Taori, Ishaan Gulrajani, Tianyi Zhang, Yann Dubois, Xuechen Li, Carlos Guestrin, Percy Liang, and Tatsunori~B. Hashimoto. 2023.
\newblock Stanford alpaca: An instruction-following llama model.
\newblock \url{https://github.com/tatsu-lab/stanford_alpaca}.

\bibitem[{Touvron et~al.(2023{\natexlab{a}})Touvron, Lavril, Izacard, Martinet, Lachaux, Lacroix, Rozi{\`e}re, Goyal, Hambro, Azhar et~al.}]{touvron2023llama}
Hugo Touvron, Thibaut Lavril, Gautier Izacard, Xavier Martinet, Marie-Anne Lachaux, Timoth{\'e}e Lacroix, Baptiste Rozi{\`e}re, Naman Goyal, Eric Hambro, Faisal Azhar, et~al. 2023{\natexlab{a}}.
\newblock Llama: Open and efficient foundation language models.
\newblock \emph{arXiv preprint arXiv:2302.13971}.

\bibitem[{Touvron et~al.(2023{\natexlab{b}})Touvron, Martin, Stone, Albert, Almahairi, Babaei, Bashlykov, Batra, Bhargava, Bhosale et~al.}]{touvron2023llama2}
Hugo Touvron, Louis Martin, Kevin Stone, Peter Albert, Amjad Almahairi, Yasmine Babaei, Nikolay Bashlykov, Soumya Batra, Prajjwal Bhargava, Shruti Bhosale, et~al. 2023{\natexlab{b}}.
\newblock Llama 2: Open foundation and fine-tuned chat models.
\newblock \emph{arXiv preprint arXiv:2307.09288}.

\bibitem[{Wan et~al.(2023)Wan, Huang, Yang, Quan, Bi, and Shi}]{wan-etal-2023-explore}
Fanqi Wan, Xinting Huang, Tao Yang, Xiaojun Quan, Wei Bi, and Shuming Shi. 2023.
\newblock \href {https://doi.org/10.18653/v1/2023.emnlp-main.587} {Explore-instruct: Enhancing domain-specific instruction coverage through active exploration}.
\newblock In \emph{Proceedings of the 2023 Conference on Empirical Methods in Natural Language Processing}, pages 9435--9454, Singapore. Association for Computational Linguistics.

\bibitem[{Wang et~al.(2023)Wang, Kordi, Mishra, Liu, Smith, Khashabi, and Hajishirzi}]{wang-etal-2023-self-instruct}
Yizhong Wang, Yeganeh Kordi, Swaroop Mishra, Alisa Liu, Noah~A. Smith, Daniel Khashabi, and Hannaneh Hajishirzi. 2023.
\newblock \href {https://doi.org/10.18653/v1/2023.acl-long.754} {Self-instruct: Aligning language models with self-generated instructions}.
\newblock In \emph{Proceedings of the 61st Annual Meeting of the Association for Computational Linguistics (Volume 1: Long Papers)}, pages 13484--13508, Toronto, Canada. Association for Computational Linguistics.

\bibitem[{Wei et~al.(2022)Wei, Bosma, Zhao, Guu, Yu, Lester, Du, Dai, and Le}]{wei2022finetuned}
Jason Wei, Maarten Bosma, Vincent Zhao, Kelvin Guu, Adams~Wei Yu, Brian Lester, Nan Du, Andrew~M. Dai, and Quoc~V Le. 2022.
\newblock \href {https://openreview.net/forum?id=gEZrGCozdqR} {Finetuned language models are zero-shot learners}.
\newblock In \emph{International Conference on Learning Representations}.

\bibitem[{Wenzek et~al.(2020)Wenzek, Lachaux, Conneau, Chaudhary, Guzm{\'a}n, Joulin, and Grave}]{wenzek-etal-2020-ccnet}
Guillaume Wenzek, Marie-Anne Lachaux, Alexis Conneau, Vishrav Chaudhary, Francisco Guzm{\'a}n, Armand Joulin, and Edouard Grave. 2020.
\newblock \href {https://aclanthology.org/2020.lrec-1.494} {{CCN}et: Extracting high quality monolingual datasets from web crawl data}.
\newblock In \emph{Proceedings of the Twelfth Language Resources and Evaluation Conference}, pages 4003--4012, Marseille, France. European Language Resources Association.

\bibitem[{Xu et~al.(2023)Xu, Sun, Zheng, Geng, Zhao, Feng, Tao, and Jiang}]{xu2023wizardlm}
Can Xu, Qingfeng Sun, Kai Zheng, Xiubo Geng, Pu~Zhao, Jiazhan Feng, Chongyang Tao, and Daxin Jiang. 2023.
\newblock Wizardlm: Empowering large language models to follow complex instructions.
\newblock \emph{arXiv preprint arXiv:2304.12244}.

\bibitem[{Xue et~al.(2021)Xue, Constant, Roberts, Kale, Al-Rfou, Siddhant, Barua, and Raffel}]{xue-etal-2021-mt5}
Linting Xue, Noah Constant, Adam Roberts, Mihir Kale, Rami Al-Rfou, Aditya Siddhant, Aditya Barua, and Colin Raffel. 2021.
\newblock \href {https://doi.org/10.18653/v1/2021.naacl-main.41} {m{T}5: A massively multilingual pre-trained text-to-text transformer}.
\newblock In \emph{Proceedings of the 2021 Conference of the North American Chapter of the Association for Computational Linguistics: Human Language Technologies}, pages 483--498, Online. Association for Computational Linguistics.

\bibitem[{Yong et~al.(2023)Yong, Menghini, and Bach}]{yong2023low}
Zheng-Xin Yong, Cristina Menghini, and Stephen~H Bach. 2023.
\newblock Low-resource languages jailbreak gpt-4.
\newblock \emph{arXiv preprint arXiv:2310.02446}.

\bibitem[{Zheng et~al.(2023)Zheng, Chiang, Sheng, Zhuang, Wu, Zhuang, Lin, Li, Li, Xing et~al.}]{zheng2023judging}
Lianmin Zheng, Wei-Lin Chiang, Ying Sheng, Siyuan Zhuang, Zhanghao Wu, Yonghao Zhuang, Zi~Lin, Zhuohan Li, Dacheng Li, Eric Xing, et~al. 2023.
\newblock Judging llm-as-a-judge with mt-bench and chatbot arena.
\newblock \emph{arXiv preprint arXiv:2306.05685}.

\bibitem[{Zhou et~al.(2023)Zhou, Liu, Xu, Iyer, Sun, Mao, Ma, Efrat, Yu, Yu et~al.}]{zhou2023lima}
Chunting Zhou, Pengfei Liu, Puxin Xu, Srini Iyer, Jiao Sun, Yuning Mao, Xuezhe Ma, Avia Efrat, Ping Yu, Lili Yu, et~al. 2023.
\newblock Lima: Less is more for alignment.
\newblock \emph{arXiv preprint arXiv:2305.11206}.

\end{thebibliography}
